# Machine Learning Emulation of Urban Land Surface Processes

David Meyer[1,2] (ORCID: 0000-0002-7071-7547)

Sue Grimmond[1] (ORCID: 0000-0002-3166-9415)

Peter Dueben[3] (ORCID: 0000-0002-4610-3326)

Robin Hogan[3,1] (ORCID: 0000-0002-3180-5157)

Maarten van Reeuwijk[2] (ORCID: 0000-0003-4840-5050)

[1]Department of Meteorology, University of Reading, Reading, UK

[2]Department of Civil and Environmental Engineering, Imperial College London, London, UK

[3]European Centre for Medium-Range Weather Forecasts, Reading, UK

Correspondence to David Meyer (d.meyer@pgr.reading.ac.uk)

**Key Points:**

- Ensemble mean of several urban land surface models is accurately emulated using a neural network emulator

- The emulator reduces computational time and complexity compared to a typical urban land surface model

- Coupled to a numerical weather model, the emulator produces accurate and stable forecasts

## Abstract

Can we improve the modeling of urban land surface processes with machine learning (ML)? A prior comparison of urban land surface models (ULSMs) found that no single model is "best" at predicting all common surface fluxes. Here, we develop an urban neural network (UNN) trained on the mean predicted fluxes from 22 ULSMs at one site. The UNN emulates the mean output of ULSMs accurately. When compared to a reference ULSM (Town Energy Balance; TEB), the UNN has greater accuracy relative to flux observations, less computational cost, and requires fewer input parameters. When coupled to the Weather Research Forecasting (WRF) model using TensorFlow bindings, WRF-UNN is stable and more accurate than the reference WRF-TEB. Although the application is currently constrained by the training data (1 site), we show a novel approach to improve the modeling of surface fluxes by combining the strengths of several ULSMs into one using ML.

## Plain Language Summary

Climate change and densely populated cities make the task of urban weather and climate prediction more and more critical to our society. In this study, we use machine learning to improve the accuracy and efficiency of models predicting urban weather. We find great potential to use these types of machine learning models both as standalone tools and integrated into complex weather models.

## 1. Introduction

Land surface models (LSM) parameterize energy exchanges between the surface and the atmosphere, providing the lower boundary conditions (e.g., radiative and turbulent heat fluxes) to atmospheric models (Stensrud, 2007). For urban areas, ULSMs (urban LSMs) are currently employed in some operational numerical weather prediction (e.g., Bengtsson et al., 2017; Seity et al., 2011) and global climate models (e.g., Hertwig et al., 2021; Oleson et al., 2011) at the higher spatial resolution end, but there is a growing need for broader adoption as they are fundamental to the delivery of integrated urban services (Baklanov et al., 2018; Grimmond et al., 2020). The complexity of ULSMs varies from simple assumptions (e.g., characterizing an impervious slab) to models that consider the 3D geometry of buildings with varying heights and material characteristics (Grimmond et al., 2009, 2010). This higher complexity, however, often comes at the cost of a greater number of site-specific input parameters and increased computational cost, which does not necessarily translate into improved results (Grimmond et al., 2011).

In recent years, machine learning (ML) techniques have shown potential in several areas of meteorology (e.g., Bolton & Zanna, 2019; Krasnopolsky et al., 2013; Nowack et al., 2018; Rasp et al., 2018; Rasp & Lerch, 2018). A key limitation of these techniques, however, is the need for large amounts of training data which, in urban meteorology, are often scarce.

One alternative to this is the creation of ML emulators (i.e., statistical surrogates of their physical counterparts) to improve the computational performance for a trade-off in accuracy (Meyer et al., 2021). Although emulators seek to improve the computational performance of current physical parameterizations, they offer no improvement in accuracy as surrogate models are, at best, as good as the data they are trained on. In urban land surface modeling, speed is, however, not such a limitation; unlike processes such as radiative transfer where the fundamental processes are well understood but computational cost is the primary limiting factor (Meyer et al., 2022), most current ULSMs are reasonably fast but require several input parameters. Furthermore, previous comparisons (Grimmond et al., 2010, 2011) found that no individual ULSM is best at predicting all the main surface fluxes such as short- and longwave radiation, and turbulent sensible and latent heat fluxes. Although an obvious solution to this issue may be mitigated by running an ensemble of ULSMs coupled to a weather (or climate) model and use it to improve predictions, this is technically challenging to implement and hard to defend given the multi-fold increase in the computational cost resulting from running multiple ULSMs at once. Moreover, given the complexity of ULSMs, their availability, and the number of specific parameters needed to make realistic simulations, ULSMs often need a specialized team of people while an ML emulator may learn the behavior of an ensemble mean and be a cheaper and easier alternative to run.

Here, we seek to develop an emulator of urban land surface processes and evaluate whether the strengths of multiple ULSMs can be combined to improve both accuracy and computational performance. The specific goals of this paper are:

1. To develop an ML emulator of urban land surface processes trained on the outputs of several ULSMs
2. To evaluate the emulator's accuracy and computational performance
3. To couple the developed ML emulator to a numerical weather model and to evaluate its accuracy and stability

To our knowledge, this is the first attempt to emulate a ULSM. In the following sections, we introduce the general problem of urban land surface modeling with details about data and methods used to develop the emulator (Section 2) and analyze the results (Section 3) before concluding with a summary and ideas for further work (Section 4).

## 2. Methods

### 2.1 General Problem

At the core of ULSMs is the concept of surface energy balance (SEB), a general statement of energy conservation with applications to surfaces and volumes of all temporal scales (Oke et al., 2017). Physically, it describes the heating (or cooling) of a surface (Figure 1). Mathematically, it can be stated as:

$$\frac{dQ_S}{dt} = Q^\star - Q_H - Q_E, \tag{1}$$

where $dQ_S/dt$ is rate of change in thermal energy stored in a surface by conduction with $Q_S$ the heat storage; $Q^\star = (S^\downarrow - S^\uparrow) + (L^\downarrow - L^\uparrow)$ is the surface net all-wave radiation flux density from downwelling ($\downarrow$) and upwelling ($\uparrow$) shortwave ($S$) and longwave ($L$) radiation; the convective heat flux densities are $Q_H$ the turbulent sensible and $Q_E$ the turbulent latent (or evaporative). Anthropogenic heat fluxes, the additional energy fluxes associated with human activities, if not simulated or prescribed, may be assumed to be zero or minimal in ULSMs (e.g., in low-density residential areas). The horizontal advection of heat and moisture is generally ignored or parameterized by ULSMs but implicitly included when coupled to weather models. ULSMs generally solve a prognostic equation in the form of Equation 1 to predict the evolution of upwelling short- and longwave radiation flux density, sensible and latent heat flux density, forced by downwelling short- and longwave radiation flux density, air temperature and humidity, atmospheric pressure, wind speed and direction and liquid (or solid) precipitation.

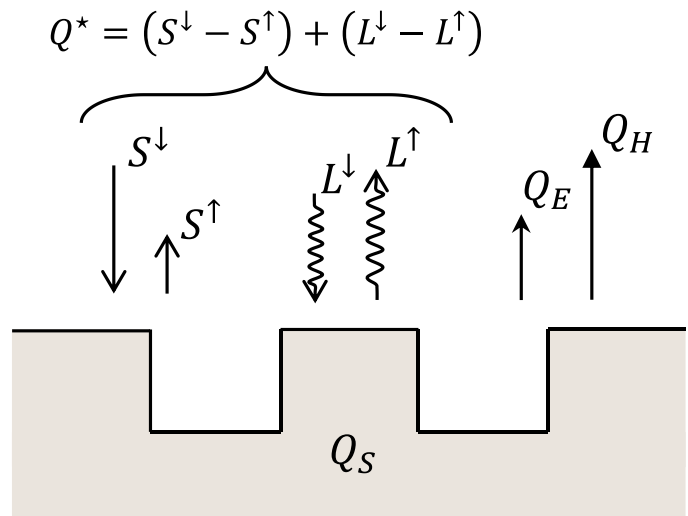

**Figure 1.** Conceptual sketch of surface energy balance exchanges (ignoring advection, vegetation, and anthropogenic heat flux) for a clear-sky day with arrows indicating the direction of fluxes relative to the surface. The shortwave flux is zero at night as the Sun's rays are below the horizon. The heat transfer processes are (left to right) radiative and convective. Other terms are defined in the text and Table 1.

## 2.2 Urban Neural Network

The urban neural network (UNN) developed here is based on the multilayer perceptron (MLP; Bishop, 2006; Goodfellow et al., 2016), one of the simplest types of neural networks (NNs). The MLP-based UNN (Figure 2) is set up to predict upwelling short- and longwave radiation flux density and turbulent sensible and latent heat flux density (Table 1b) at time $t+1$ from inputs at time $t$ of common meteorological variables such as dry-bulb air temperature and humidity, as well as the cosine of solar zenith angle $\mu_0$ and model timestep length $\Delta t$ (Table 1a). Both $\mu_0$ and $\Delta t$ are used to allow the UNN to run over different grid points at different spatial and temporal resolution when coupled to the weather model (section 2.4). Specifically, $\mu_0$ is used instead of latitude, longitude, and local time to reduce the number of features required by the UNN, and $\Delta t$ to dynamically vary the timestep length, matching that used by the weather model. The surface temperature $T_s$ is used rather than the upwelling longwave radiation $L^\uparrow$ to mimic a physical system whereby $T_s$ is used as a state between different timesteps and thus provide the initial condition at each new inference timestep (Figure 2). The UNN is implemented in TensorFlow (Abadi et al., 2015) version 2.6.2 (TensorFlow Developers, 2021c) and configured with two hidden layers, each having 256 neurons, rectified linear unit (ReLU) activation function, and Adam optimizer (Kingma & Ba, 2015) with mean squared error as its optimization function. This configuration is deemed optimal after conducting a hyperparameter optimization of several configurations (Table S1) and visually inspecting the results.

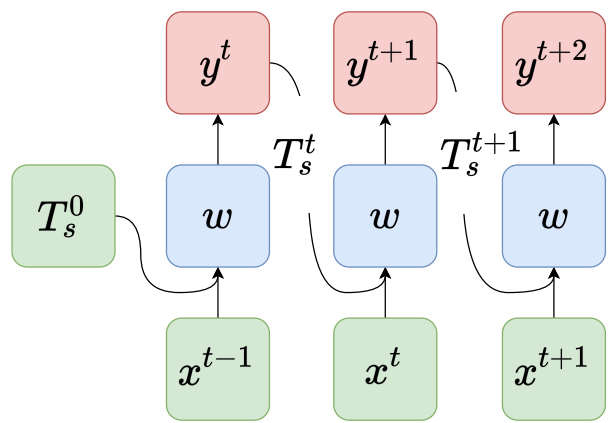

**Figure 2**. The trained urban neural network (UNN) is used to predict outputs **y** (Table 1b) at time $t+1$ from inputs **x** (Table 1a) at time $t$ using trained weights **w**. The surface temperature $T_s$ from a previous timestep provides the initial condition at each new inference.

**Table 1**. Inputs and outputs from data sets and models used to conduct the simulations. Multi-model Ensemble Mean (MEM). Depending on the model or data set used, data may be unavailable/not-applicable (-), available/outputted (✓), derived (D), constant (C), unmodified (U), state (S). †Additional inputs are given in namelists. ᵃTEB requires direct and diffuse; these are computed using pvlib (Holmgren et al., 2018) version 0.9.0 (Holmgren et al., 2021) but given directly as-is in WRF. ᵇWind speed and direction are used instead of, and derived from the zonal and meridional components of wind velocity using MetPy version 1.1.0 (May et al., 2021). For computations between upwelling longwave radiation flux density $L^\uparrow$ and surface temerature $T_s$ a constant emissivity ($\varepsilon$) of 0.97 as reported in Coutts et al. (2007b) is used. The latent heat of vaporization $\mathcal{L}_v$ is assumed to be constant with a value of 2.464 MJ kg⁻¹ which is applicable to 15 °C (Oke et al., 2017).

| Symbol | Name | Unit | MEM | TEB† | WRF-TEB† | UNN | WRF-UNN | Derived as |
|---|---|---|---|---|---|---|---|---|
| *(a) Inputs* | | | | | | | | |
| $T$ | Dry-bulb air temperature | K | ✓ | ✓ | ✓ | ✓ | ✓ | - |
| $q$ | Specific humidity | kg kg⁻¹ | ✓ | ✓ | ✓ | ✓ | ✓ | - |
| $p$ | Atmospheric surface pressure | Pa | ✓ | ✓ | ✓ | ✓ | ✓ | - |
| $S^\downarrow$ | Downwelling shortwave radiation flux density | W m⁻² | ✓ | ✓ᵃ | ✓ᵃ | ✓ | ✓ | - |
| $L^\downarrow$ | Downwelling longwave radiation flux density | W m⁻² | ✓ | ✓ | ✓ | ✓ | ✓ | - |
| $u$ | Zonal component of wind velocity | m s⁻¹ | ✓ᵇ | ✓ᵇ | ✓ | ✓ | ✓ | - |
| $v$ | Meridional component of wind velocity | m s⁻¹ | ✓ᵇ | ✓ᵇ | ✓ | ✓ | ✓ | - |
| RR | Rainfall rate | kg m⁻² s⁻¹ | ✓ | ✓ | ✓ | ✓ | ✓ | - |
| $t_{local}$ | Local time | s | ✓ | ✓ | ✓ | - | - | - |
| $\varphi$ | Latitude | deg | ✓ | ✓ | ✓ | - | - | - |
| $\lambda$ | Longitude | deg | ✓ | ✓ | ✓ | - | - | - |
| $\mu_0$ | Cosine of solar zenith angle | rad | - | - | - | ✓ᵃ | ✓ | - |
| $\Delta t$ | Timestep length | s | ✓ | ✓ | ✓ | ✓ | ✓ | - |
| *(b) Outputs* | | | | | | | | |
| $S^\uparrow$ | Upwelling shortwave radiation flux density | W m⁻² | ✓ | ✓ | - | ✓ | ✓ | - |
| $L^\uparrow$ | Upwelling longwave radiation flux density | W m⁻² | ✓ | ✓ | - | - | D | $\varepsilon\sigma T_s^4$ |
| $T_s$ | Surface (skin) temperature | K | - | ✓ | ✓ | S | S | $[L^\uparrow / (\varepsilon\sigma)]^{1/4}$ |
| $Q_H$ | Turbulent sensible heat flux density | W m⁻² | ✓ | ✓ | ✓ | ✓ | ✓ | - |
| $Q_E$ | Turbulent latent heat flux density | W m⁻² | ✓ | ✓ | ✓ | ✓ | ✓ | - |
| $E$ | Evaporation mass flux density | kg m⁻² s⁻¹ | - | ✓ | ✓ | - | D | $Q_E / \mathcal{L}_v$ |
| $Q_S$ | Heat Storage | J m⁻² | - | ✓ | ✓ | - | D | Equation 1 |
| $\alpha$ | Surface albedo | 1 | C | C | C | - | D | $S^\uparrow/S^\downarrow$ |
| $\varepsilon$ | Surface emissivity | 1 | C | C | C | - | - | - |
| $w_s$ | Mass mixing ratio of water vapor | kg kg⁻¹ | - | ✓ | ✓ | - | U | - |
| $u_*$ | Shear (friction) velocity | m s⁻¹ | - | ✓ | ✓ | - | U | - |

## 2.3 Town Energy Balance

To compare the UNN to a baseline, here we use the Town Energy Balance (TEB; Masson, 2000) model, a single-layer ULSM characterizing city areas based on building roofs, walls, roads, and vegetation, and assuming buildings create an infinite street canyon (Masson, 2000). TEB is chosen as it is a mature, widely used ULSM, extensively evaluated (e.g., Lemonsu et al., 2004; Masson et al., 2002; Pigeon et al., 2008), and available both offline (Meyer, Schoetter, Masson, et al., 2020) and online (i.e., coupled to weather models; e.g., Hamdi et al., 2012; Lemonsu & Masson, 2002; Meyer, Schoetter, Riechert, et al., 2020). Here we use the TEB software (Meyer, Schoetter, Masson, et al., 2020) version 4.1.2 (Masson et al., 2021) and refer to it as TEB. Similar to the UNN, TEB inputs include typical meteorological variables at time $t$ such as dry-bulb air temperature and humidity (Table 1a) to predict common surface energy balance variables at time $t + 1$ (Table 1b).

## 2.4 Weather Research and Forecasting (WRF) Coupling

Both TEB and the UNN are coupled to the weather model WRF (Weather Research and Forecasting; Skamarock et al., 2019) in WRF-CMake (Riechert & Meyer, 2019b) version 4.2.2 (Riechert & Meyer, 2021) as it simplifies WRF-related development, configuration, and build-processes. Implementation details of WRF-TEB are provided in Meyer, Schoetter, Riechert, et al. (2020). Variables used in the WRF-UNN and WRF-TEB coupling are similar; however, as friction velocity $u_*$ data are not provided in either observations or most ULSMs (Figure S1), it is not part of the UNN (Section 2.2) and thus ignored in WRF-UNN.

Porting the UNN to WRF is made seamless by relying on the available C application programming interface (API) provided with TensorFlow (TensorFlow Developers, 2021a). We choose the lightweight version of TensorFlow, TensorFlow Lite (TensorFlow Developers, 2021b) as: (a) it has CMake support, which makes the integration in WRF-CMake straightforward and allows sharing of project build options in WRF-CMake, and (b) it has a very succinct and accessible API (compared to TensorFlow library C API), that makes the Fortran binding used in the coupling easy to write. To perform the actual coupling, the UNN is exported to a TensorFlow Lite file from Python. To enable the UNN in WRF, the TFLite Fortran binding (`tflite.f90`; Figure 3a) and the UNN surface module (`module_sf_unn.F`; Figure 3) are written. The former is used to interface with the TFLite C API and the latter to initialize inputs (Table 1a) and run the UNN to generate outputs (Table 1) which are passed to WRF for the next timestep.

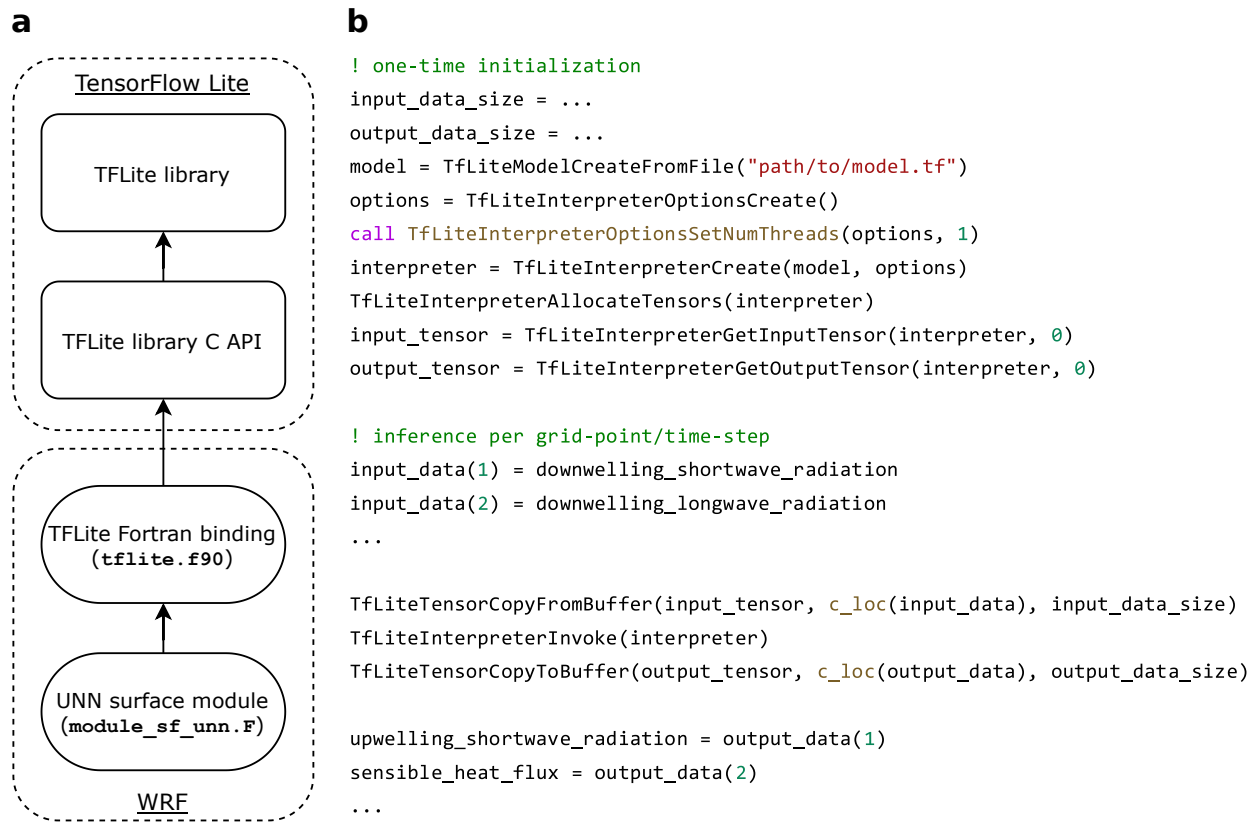

**a**

**TensorFlow Lite**

TFLite library

TFLite library C API

TFLite Fortran binding
(`tflite.f90`)

UNN surface module
(`module_sf_unn.F`)

WRF

**b**

```
! one-time initialization
input_data_size = ...
output_data_size = ...
model = TfLiteModelCreateFromFile("path/to/model.tf")
options = TfLiteInterpreterOptionsCreate()
call TfLiteInterpreterOptionsSetNumThreads(options, 1)
interpreter = TfLiteInterpreterCreate(model, options)
TfLiteInterpreterAllocateTensors(interpreter)
input_tensor = TfLiteInterpreterGetInputTensor(interpreter, 0)
output_tensor = TfLiteInterpreterGetOutputTensor(interpreter, 0)

! inference per grid-point/time-step
input_data(1) = downwelling_shortwave_radiation
input_data(2) = downwelling_longwave_radiation
...

TfLiteTensorCopyFromBuffer(input_tensor, c_loc(input_data), input_data_size)
TfLiteInterpreterInvoke(interpreter)
TfLiteTensorCopyToBuffer(output_tensor, c_loc(output_data), output_data_size)

upwelling_shortwave_radiation = output_data(1)
sensible_heat_flux = output_data(2)
...
```

**Figure 3.** Call to Tensorflow Lite (TFLite) C API from WRF as implemented for this study. (**a**) TFLite integration: the Fortran interface module `tflite.f90` is used in WRF to bind to the TFLite C API, which is used by the UNN surface module `module_sf_unn.F` to initialize and run the neural network. (**b**) Example of TFLite Fortran binding (simplified from UNN surface module) use: first, a one-time initialization to load the TFLite model, configure its settings, and allocate input and output tensors is performed. Before each inference, input quantities (e.g., downwelling shortwave radiation) are stored in the input tensor in the expected order. Similarly, inference outputs are returned as output tensors from which individual quantities are accessed. Other pre- and post-processing steps such as normalization and error handling are omitted. See `models/wrf-unn/phys/module_sf_unn.F` and `models/wrf-unn/phys/module_sf_unn.F` in Meyer (2021) for the actual code.

### 2.5 Data and Model Setup

Grimmond et al. (2011)'s urban comparison study evaluated the accuracy of 32 ULSMs (or different configurations) from a wide range of international modeling groups (Table S2) using directly observed fluxes in Preston, a suburban area of Melbourne (Australia). Site information to configure ULSMs, released to participants in four *Stages*, gives increasing detail (Table S3). For each Stage, groups returned their model computed surface energy balance fluxes. The main data sets, with local time stamps (UTC+10, i.e., 10-h ahead of Coordinated Universal Time), are:

1. *Morphological parameters* (MP): provided in the four Stages, characterizing the surface around the observation site (Grimmond et al., 2011 Table 2, and our Table S3).

2. *Meteorological Forcing* (MF; Grimmond et al., 2021): continuous gap-filled 30-min averages with a period ending timestamp (i.e., 10:30 is 10:01–10:30) of meteorological variables at 40 m above ground level (agl) between 12 August 2003 13:30 and 28 November 2004 23:00 (Figure 4; Table 1a).

3. *Multi-model output* (MO; Grimmond et al., 2013): continuous outputs, from 32 models or model-configurations between 13 August 2003 00:00 and 27 November 2004 23:30 reported in four separate Stages for upwelling short- and longwave radiation flux density, and turbulent sensible and latent heat flux density (Table 1b).

4. *Observations* (OBS; Grimmond et al., 2021): 30-min average fluxes with period ending timestamp measured at 40 m agl between 13 August 2003 00:00 and 27 November 2004 23:30. Methods to obtain observed fluxes are given in in Coutts et al. (2007a, 2007b). This data set has the same fluxes as the MO data set but with observational gaps (~39% of MO; purple Figure 4a).

Here, a visual inspection of the MO data set for all Stages (Figure S2) is used to remove ULSMs not simulating the turbulent latent heat flux density (8 ULSMs) or showing outliers (2 ULSMs), leaving 22 ULSMs (Figure S3). The MO data set is used to compute the ensemble mean, hereafter referred to as the multi-model ensemble mean (MEM). To make evaluations consistent between MEM, UNN, and TEB, these are conducted for periods spanning 13 August 2003 00:00 and 27 November 2004 23:30 as defined in MEM totaling 22 704 30-min samples. For any given data set, the test fraction is the periods with OBS available (i.e., 39 % of MEM; number of samples $N$ = 8 866; purple, Figure 4a) and the training fraction (used by the UNN; see point 2 below) for the remaining periods (i.e., 61 % of MEM; $N$ = 13 838; black, Figure 4a). Evaluation metrics (Section 2.6) are calculated for the test fraction, with results (Section 3) using all samples except for the upwelling shortwave radiation flux density as this is zero at nighttime (i.e., daytime: OBS > 2 W m$^{-2}$, $N$ = 4 272).

Model-specific setups are as follows:

1. TEB uses morphological parameters from MP for the four Stages (Table S3) forced with MF. TEB is run with 5-min (300-s) timesteps (from 13 August 2003 00:00 to 27 November 2004 23:30) after linear interpolation of the 30-min MF data set (e.g., 00:00, 00:05, …) to predict the next 5-min (e.g., 00:05, 00:10, …). The last 5-min sample of each 30-min period (e.g., 00:30) is used in analyses (Section 3). From the evaluation of TEB outputs at all Stages, Stage 4 is selected as it has the smallest errors (Appendix A).

2. The UNN is trained with MF as inputs and MEM from Stage 2 as outputs (Table 1) using the training fraction. Stage 2 is selected as it offers the 'best' tradeoff between complexity (i.e., number of parameters used to configure the 22 ULSMs; Table S3) and accuracy (Appendix A). Prior to training, the surface temperature $T_s$ is derived from the MEM upwelling longwave radiation flux density $L^{\uparrow}$ assuming a constant emissivity (Table 1b). To allow the UNN to be used with different timestep lengths, nine linearly interpolated copies of both inputs and outputs are made (with 1, 2, 5, 10, 20, 60, 120, 300, and 600-s timesteps), each derived from the 30-min data, and concatenated together with the original. A random subset corresponding to the same number of samples included with the 30-min data ($N$ = 13 838) is selected in each copy to keep the number of training samples across the linearly interpolated copies equal. Thus, the total number of samples used for training the UNN is 138 380 (i.e., ten times the original 30-min data). Of this, 25 % are randomly reserved for the early stopping mechanism. For inference, the UNN is forced with 5-min MF (12 August 2003 23:30 and 27 November 2004 23:30) derived by linearly interpolating the 30-min intervals to be consistent with TEB. As UNN outputs (Table 1b) include $T_s$ rather than $L^{\uparrow}$ used in evaluations, UNN outputs are postprocessed to derive $L^{\uparrow}$ from $T_s$ assuming a constant emissivity (Table 1b). The stochastic nature of the multilayer perceptrons is assessed by repeating the training (and inference) 100 times, each with a different random seed. At each iteration (a) for each UNN output variable ($S^{\uparrow}$, $L^{\uparrow}$, $Q_H$, $Q_E$; Table 1b), the normalized mean absolute error (nMAE; section 2.6) is computed using the 'true' MEM and UNN-predicted samples for the whole period (i.e., both train and test fractions) and (b) the mean nMAE ($\overline{\text{nMAE}}$) is computed as $0.25 \left( \text{nMAE}_{S^\uparrow} + \text{nMAE}_{L^\uparrow} + \text{nMAE}_{Q_H} + \text{nMAE}_{Q_E} \right)$. The UNN with the median $\overline{\text{nMAE}}$ from the 100 iterations (Figure S4) is taken as the representative UNN and used in analyses (Section 3). Thus, all UNN-relevant metrics (Section 2.6) are computed using results from the UNN with the median $\overline{\text{nMAE}}$.

3. The coupled WRF-TEB and WRF-UNN simulations are set up with four nested domains (Figure 5), generated with GIS4WRF (Meyer & Riechert, 2019) version 0.14.4 (Meyer & Riechert, 2020) and processed using WPS-CMake (WRF Preprocessing System) version 4.1.0 (Riechert & Meyer, 2019a). Both TEB and the UNN are run for the innermost domain (Figure 5b) with a 5-s timestep centered on the Preston measurement tower. The innermost domain with 1 km horizontal grid spacing has a 66 m vertical grid spacing close to the surface, increasing with height. As model buildings are assumed to be within the ground (Meyer, Schoetter, Riechert, et al., 2020) the flux tower sensors at 40 m agl is 33.6 m above the model surface (as mean building height = 6.4 m, Table S3). WPS MODIS land use data (UCAR, 2019) are used as one urban class with the same urban and vegetation fractions (i.e., all grid cells) for consistency between TEB and UNN simulations. Stage 4 parameters (Table S3) are used in both WRF-TEB, and TEB-offline runs. The European Centre for Medium-Range Weather Forecasts (ECMWF) Cycle 28r2 analysis (ECMWF, 2004) are used to provide the initial and boundary conditions. Other parameters used to configure WRF and WPS are given in Table S4. Simulations are run for summer (23 December 2003 10:00–27 December 2003 10:00) and winter (25 June 2004 10:00–31 June 2004 04:00) periods, with evaluations using 65 hr in summer (24 December 2003 14:30–27 December 2003 07:30) and 98.5 hr in winter (26 June 2004 21:00–30 June 2004 23:30) to allow some model spin-up; giving the longest continuous observation evaluation periods in the two seasons. Instantaneous WRF fluxes at each 5-min interval (e.g., 00:00, 00:05, …) are averaged to 30-min time ending values for comparison with OBS.

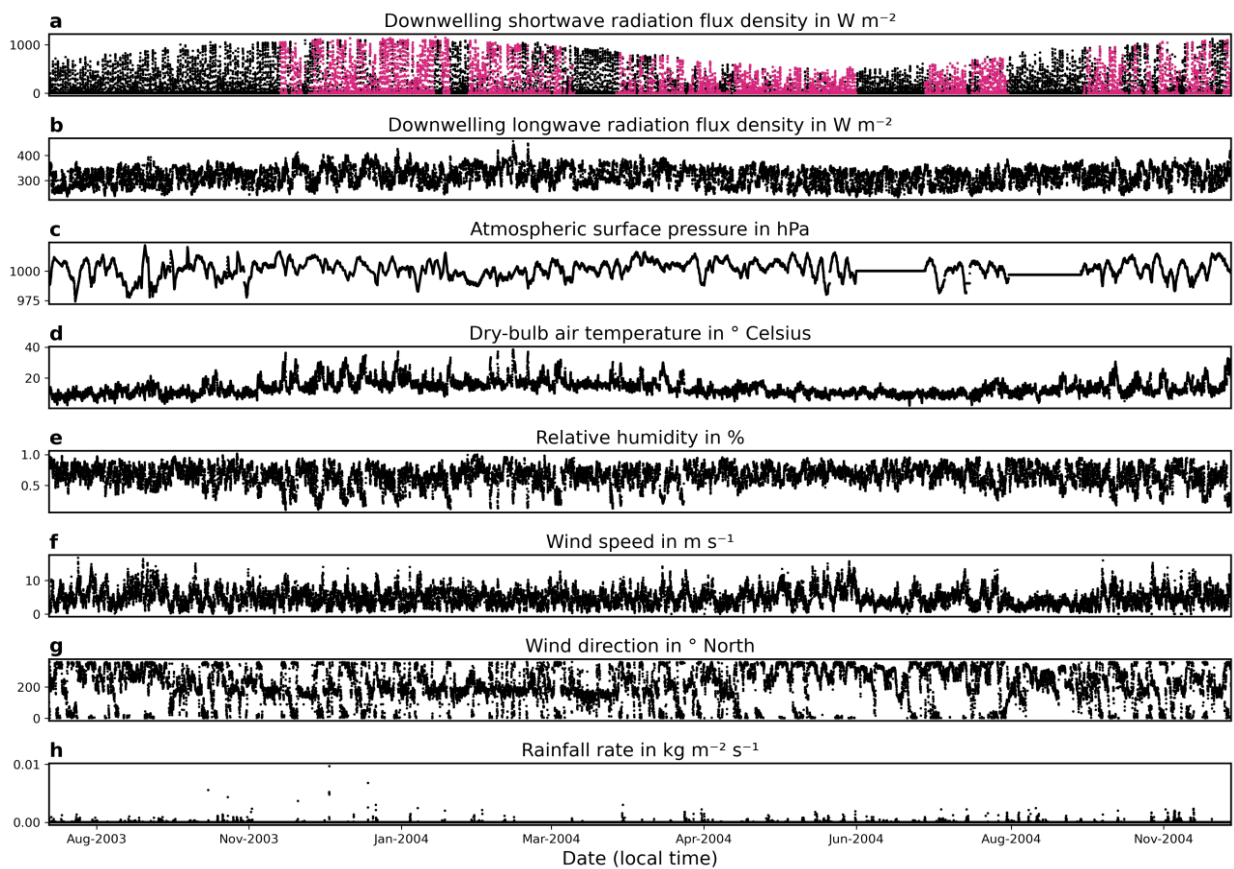

**Figure 4.** Observed meteorological forcing (MF) data (30-min, sources section 2.5): (**a**) downwelling shortwave radiation flux density with period with evaluation observed fluxes (i.e., test fraction) shown (purple), (**b**) downwelling longwave radiation flux density, (**c**) atmospheric surface pressure, (**d**) dry-bulb air temperature, (**e**) relative humidity, (**f**) wind speed and (**g**) wind direction, and (**h**) rainfall rate. Wind speed and direction computed with MetPy version 1.1.0 (May et al., 2021) from the zonal and meridional components of wind velocity. Relative humidity computed with PsychroLib (Meyer & Thevenard, 2019) version 2.5.0 (Meyer & Thevenard, 2020) from dry-bulb air temperature, specific humidity, and atmospheric surface pressure. Local time is 10-h ahead of Coordinated Universal Time (UTC+10).

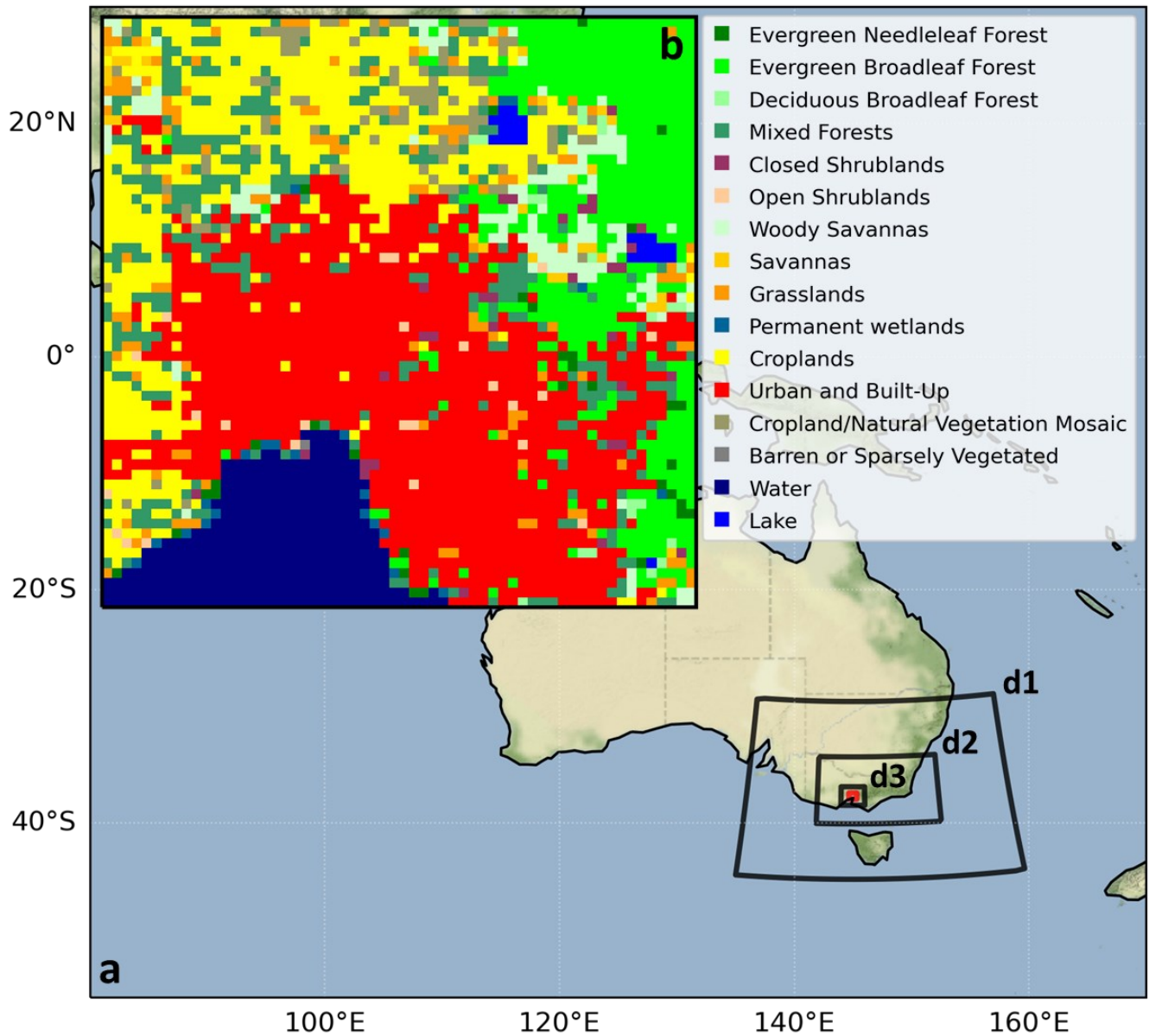

**Figure 5**. Study area (**a**) within Australia and the four nested domains (d1-d3 and innermost, red) used in online simulations, and (**b**) innermost domain (1 km horizontal grid spacing) with WPS MODIS 30 arc-sec land cover/use (UCAR, 2019). The UNN and TEB are run for 'urban' (red) grid cells in WRF assuming land cover fractions of 0.445 building, 0.38 vegetation, and remainder non-building impervious. Sources: map features by Natural Earth Vector (Kelso & Patterson, 2009) are in the public domain. Map tiles by Stamen Design (2021), under Creative Commons Attribution 3.0 license (CC BY 3.0). Data from OpenStreetMap (OpenStreetMap contributors, 2017), under Open Data Commons Open Database License (ODbL).

## 2.6 Evaluation Metrics

To assess the simulations, statistics are computed between '*true*' $y^t$ and *predicted* $\hat{y}^t$ samples at time $t$ for $N$ timesteps. The metrics used are: mean bias (MB = $\frac{1}{N}\sum_{t=1}^{N}\hat{y}^t - y^t$), mean absolute error (MAE = $\frac{1}{N}\sum_{t=1}^{N}|\hat{y}^t - y^t|$), mean absolute error normalized by mean absolute 'true' flux $\overline{|y|}$ (hereafter referred to as the normalized mean absolute error, nMAE = 100 % [MAE / $\overline{|y|}$]) and standard deviation of the error (SDE = $\sqrt{\frac{1}{N}\sum_{t=1}^{N}\left[(\hat{y}^t - y^t) - \overline{(\hat{y} - y)}\right]^2}$). Depending on the evaluation type, 'true' samples $y^t$ are from either OBS or MEM and predicted samples $\hat{y}^t$ are from either MEM, UNN or TEB outputs (Section 2.5).

## 3. Results and Discussion

### 3.1 Multi-model Ensemble Mean

First, we assess the trained urban neutral network (UNN) using the test fraction (Section 2.5) of the multi-model ensemble mean (MEM) data set to determine if the UNN captures the main processes in predicting the surface energy balance. A perfect emulator would have all points on the line $x = y$ (Figure 6). The UNN has the highest skill for the daytime upwelling shortwave radiation flux density (Figure 6a): mean bias (MB) is 3.0 W m$^{-2}$, standard deviation of the error (SDE) 4.4 W m$^{-2}$, mean absolute error (MAE) 4.2 W m$^{-2}$, and normalized mean absolute error (nMAE) 7.0 %. The UNN-predicted longwave flux (Figure 6b) is slightly poorer (MB\SDE\MAE are −4.5\6.6\6.4 W m$^{-2}$) but with a lower nMAE (1.6 %) because of the larger absolute fluxes. The turbulent latent (Figure 6d) heat flux density is more accurately predicted (MB\SDE\MAE are 5.9\13.7\8.6 W m$^{-2}$ and nMAE is 32.7 %) than the sensible (MB\SDE\MAE are −6.2\21.4\16.1 W m$^{-2}$ and nMAE is 34.2 %; Figure 6c) but with larger outliers. This relative ranking is consistent with the extensive ULSM evaluation literature.

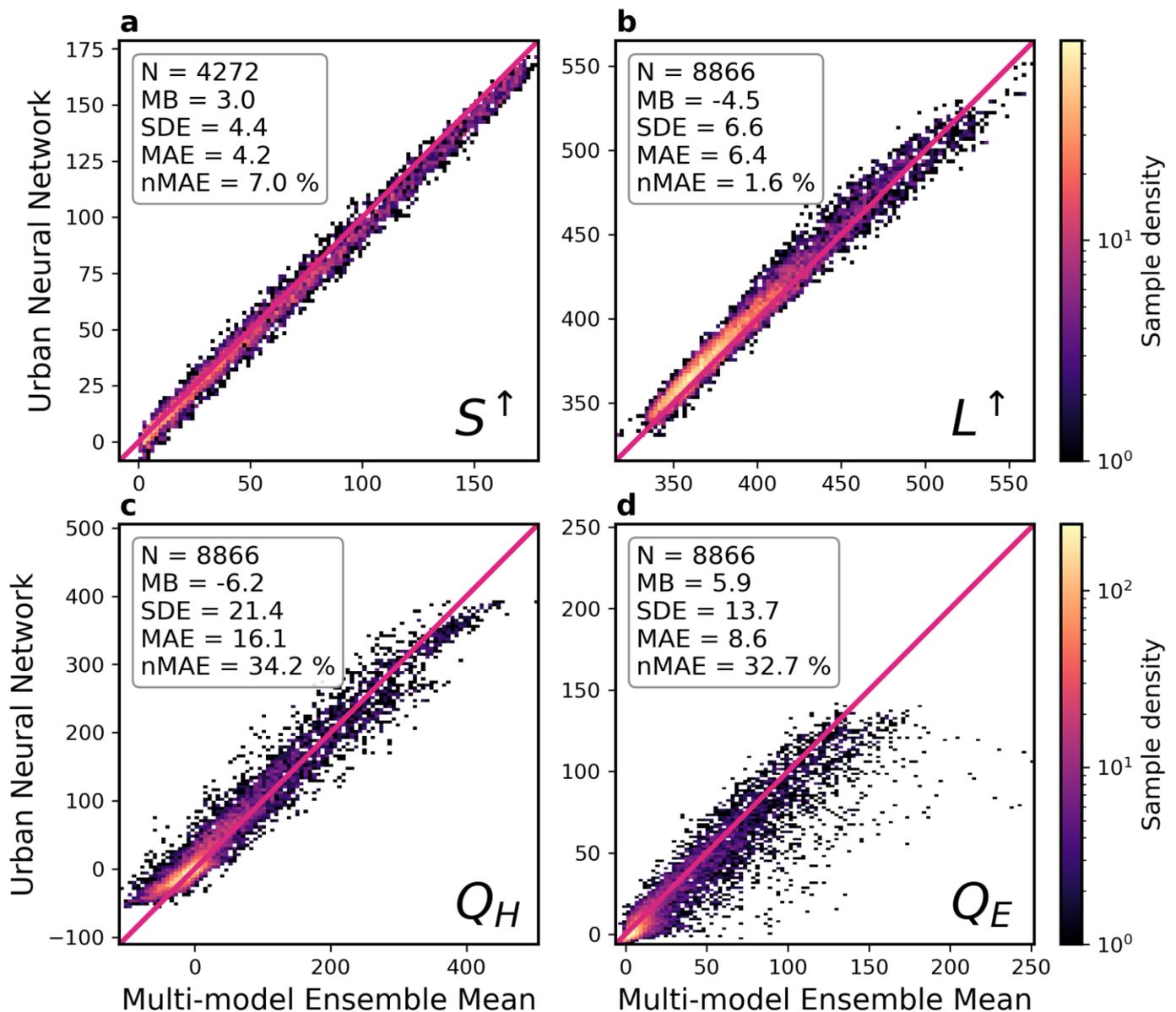

**Figure 6**. Comparison of UNN to the MEM data, with 1:1 line (red), data density (color) and evaluation statistics (section 2.6) shown, for 30-min flux densities of (**a**) daytime upwelling shortwave radiation ($S^\uparrow$) and (**b**) 24 h upwelling longwave radiation ($L^\uparrow$), turbulent (**c**) sensible ($Q_H$), and (**d**) latent ($Q_E$) heat. Units are W m$^{-2}$ except for the nMAE percentage (%). Note axes scales differ between plots.

### 3.2 Offline Simulations

Second, we compare the UNN to observed fluxes (OBS) and a version of the widely used ULSM TEB (Section 2.3). Two evaluations are undertaken with each using different meteorological forcing data: (a) observed MF (offline) and (b) coupled to WRF (online). The offline results are analyzed for both the test fraction and the short summer and winter periods (Section 2.5; Table 2). Online simulations (Section 3.3) are only evaluated for the latter two periods.

Relative to OBS, MEM has the lowest biases and errors for all but the upwelling shortwave radiation flux density, where it is outperformed by TEB (Figure 7; Table 2). Similarly, as the UNN captures the main surface processes modeled by the MEM (Section 3.1), its predictions outperform TEB's for all but the upwelling shortwave radiation. The MAE for the upwelling shortwave radiation (Figure 7a; Table 2) for all (MEM\UNN\TEB) is < 6 W m$^{-2}$ (Figures 7e–7g; Table 2). The MAE for the longwave (Figure 7b) is larger than for the shortwave, with TEB (20 W m$^{-2}$) having the largest between both MEM and UNN (MEM 4 W m$^{-2}$; UNN 7.8 W m$^{-2}$). Similarly, the MAE for the turbulent sensible (Figures 7c and 7f) and latent (Figures 7d and 7f) heat flux densities are larger for TEB (30 and 25 W m$^{-2}$, respectively) compared to MEM and UNN (≤ 21 W m$^{-2}$, Table 2). The nMAE for the latent is larger than sensible because of its smaller mean (Table 2; Figure 7g). The short summer and winter focal periods are consistent with the 16-month results (Table 2, Figure 8), with generally higher biases and errors for TEB than for either MEM or UNN. However, both MEM and UNN outperform TEB in the winter, notably with better accuracy for upwelling shortwave radiation (Figures 8b and 8i–8n; Table 2).

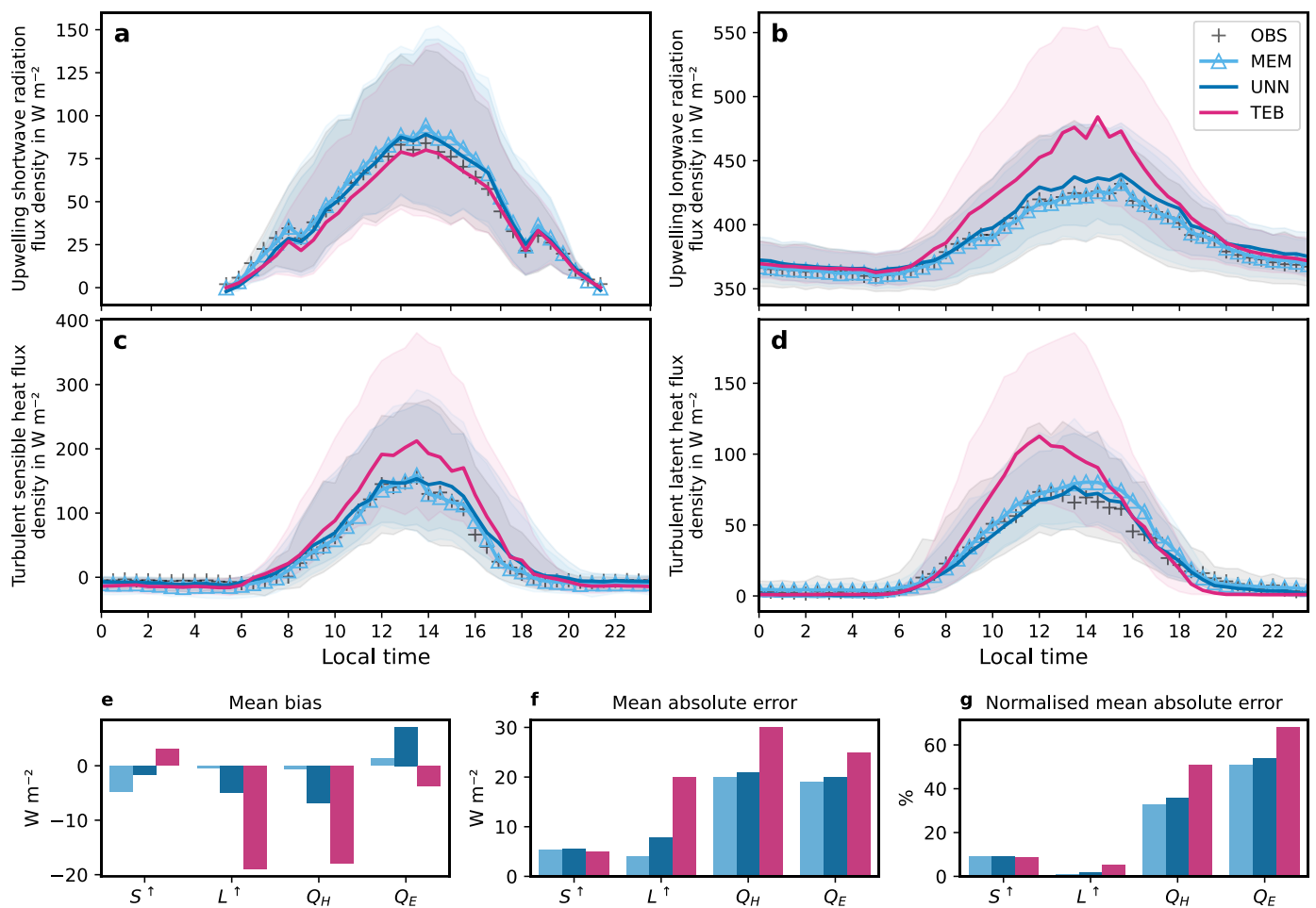

**Figure 7**. Offline (16-months. N = 8 865) simulated (MEM, UNN, TEB, see text) and observed (OBS) 30-min flux densities (lines) with interquartile range (shading) for (**a**) upwelling short- ($S^\uparrow$) (daytime) and (**b**) long-wave ($L^\uparrow$) radiation, turbulent (**c**) sensible ($Q_H$) and (**d**) latent ($Q_E$) heat, with (**e**–**g**) their respective evaluation metrics (Table 2). Note y axes differ between plots.

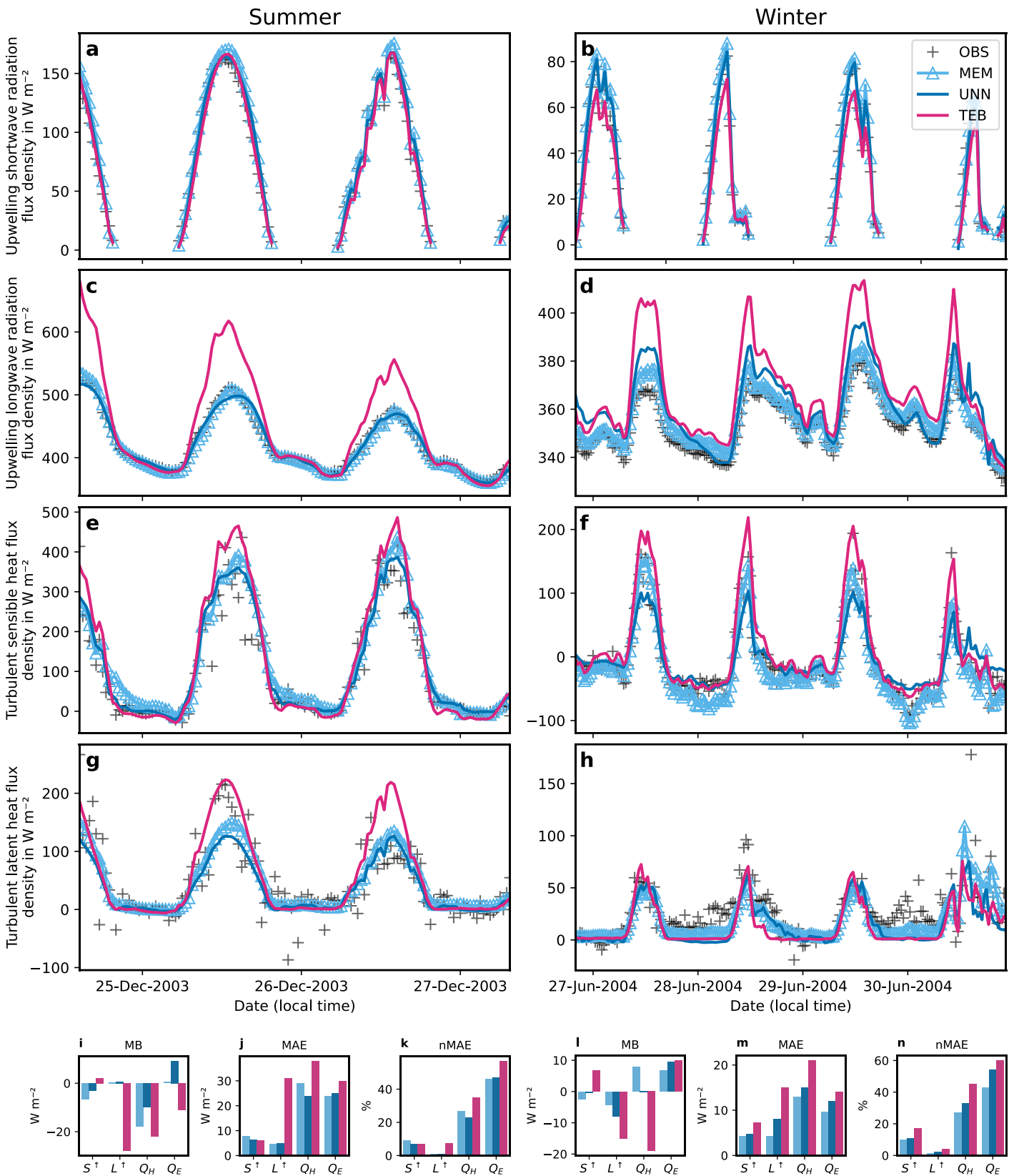

**Figure 8.** Offline simulated (MEM, UNN and TEB) and observed (OBS) 30-min flux densities in (**a,c,e,g, i-k**) a summer and (**b,d,f,h, l-n**) winter period for (**a-b**) upwelling short- ($S^{\uparrow}$) and (**c-d**) long-wave ($L^{\uparrow}$) radiation, turbulent (**e-f**) sensible ($Q_H$) and (**g-h**) latent ($Q_E$) heat, with (**i-n**) evaluation metrics (Table 2). Note y axes differ between plots.

**Table 2**. Observed flux densities (upwelling shortwave $S^\uparrow$ and longwave $L^\uparrow$ radiation, and turbulent sensible $Q_H$ and latent $Q_E$ heat) and modeled (UNN, MEM, and TEB) evaluation metrics (section 2.6) for N 30-min periods with the lowest nMAE (bold) per flux indicated per data cohort.

| | $S^\uparrow$ | $L^\uparrow$ | $Q_H$ | $Q_E$ | $S^\uparrow$ | $L^\uparrow$ | $Q_H$ | $Q_E$ | $S^\uparrow$ | $L^\uparrow$ | $Q_H$ | $Q_E$ |
|---|---|---|---|---|---|---|---|---|---|---|---|---|
| **Observed** | Mean (W m⁻²) | | | | | | | | | | | |
| _16 months_ | 58 | 390 | 40 | 34 | - | - | - | - | - | - | - | - |
| _65.5 hours in summer_ | 87 | 420 | 100 | 46 | - | - | - | - | - | - | - | - |
| _82 hours in winter_ | 41 | 350 | 22 | | - | - | - | - | - | - | - | - |
| **Simulated** | MB (W m⁻²) | | | | MAE (W m⁻²) | | | | nMAE (%) | | | |
| _16 months (13 Aug. 2003 00:00 – 27 Nov. 2004 23:30; N = 8 866, except $S^\uparrow$ N = 4 272)_ | | | | | | | | | | | | |
| MEM | -4.7 | -0.39 | -0.67 | 1.3 | 5.3 | 4.0 | 20 | 19 | 9.2 | **1.0** | 33 | 51 |
| UNN | -1.7 | -4.9 | -6.9 | 7.1 | 5.5 | 7.8 | 21 | 20 | 9.4 | 2.0 | 36 | 54 |
| TEB | 3.0 | -19 | -18 | -3.8 | 5.1 | 20 | 30 | 25 | **8.7** | 5.2 | 51 | 68 |
| _65.5 hours in summer (24 Dec. 2003 14:30 – 27 Dec. 2003 07:30; N = 131, except $S^\uparrow$ N = 73)_ | | | | | | | | | | | | |
| _Offline_: MEM | -6.6 | 0.4 | -18 | 0.53 | 7.8 | 4.6 | 29 | 24 | 9.0 | **1.1** | 27 | **46** |
| UNN | -3.2 | 0.65 | -9.8 | 9.3 | 6.3 | 5 | 24 | 25 | 7.2 | 1.2 | **23** | 47 |
| TEB | 2.1 | -28 | -22 | -11 | 6 | 31 | 38 | 30 | **6.9** | 7.3 | 35 | 57 |
| _Online_: WRF-UNN | -5.8 | 14 | 15 | -4.7 | 9.5 | 18 | 38 | 26 | 11 | **4.2** | 35 | 49 |
| WRF-TEB | -2.1 | -11 | -41 | -12 | 8.3 | 35 | 53 | 34 | **9.6** | 8.2 | 49 | 64 |
| _82 hours in winter (26 June 2004 21:00 – 30 June 2004 23:30; N = 164, except $S^\uparrow$ N = 56)_ | | | | | | | | | | | | |
| _Offline_: MEM | -2.4 | -4.3 | 7.9 | 6.8 | 4.2 | 4.3 | 13 | 9.7 | 10 | **1.2** | 27 | 43 |
| UNN | -0.43 | -7.9 | -0.24 | 9.4 | 4.7 | 8.0 | 15 | 12 | 11 | 2.3 | 34 | 54 |
| TEB | 6.7 | -15 | -19 | 10 | 7.3 | 15 | 21 | 14 | 17 | 4.2 | 45 | 60 |
| _Online_: WRF-UNN | -3.3 | -1.3 | 0.53 | 5.9 | 8.8 | 9.3 | 18 | 13 | 21 | **2.6** | 38 | 56 |
| WRF-TEB | 3.0 | -3.5 | -21 | 8.8 | 11 | 16 | 27 | 18 | 28 | 4.4 | 59 | 81 |

### 3.3 Online Simulations

The online coupled WRF-UNN and WRF-TEB simulation results for the grid cell centered on the measurement site are shown in Figure 9. The numerical stability of WRF-UNN (trained using MEM for Stage 2; Section 2.5) is demonstrated by executing thousands of iterations for hundreds of grid cells without numerical failure (i.e., 2-week, hundreds of domain grid points). The WRF-UNN post-spin-up periods for both winter and summer (Figure 9) capture the general trends of observations. Despite being derived using data from simulations driven with Stage 2 parameters (Table S3), WRF-UNN has better predictive skills than WRF-TEB (using Stage 4 parameters) for both summer and winter periods (Figure 9). WRF-UNN errors are generally lower than those of WRF-TEB in both seasons for all but the winter upwelling shortwave radiation flux density (Figure 9; Table 2) and generally consistent with the errors shown for offline simulations (Section 3.2).

As online simulations are run for the entire domain, an additional, albeit qualitative, comparison can be made across the spatial domain. The inner domain surface cover is assigned the same land cover fractions (building 0.445, paved/road 0.175, vegetated 0.380) for all 'urban' grid cells (d4 red, Figure 5) in both WRF-TEB and WRF-UNN. Given the temporal difference between the two simulations is greatest around midday on 25 December 2003 (Figure 9c), we select this time for the spatial comparisons (Figure 10). Although the simulated upwelling shortwave radiation flux density (Figures 10a and 10b) has a similar pattern across the domain, the longwave (Figures 10c and 10d) in WRF-UNN has a smaller magnitude and spatial range across all the urban grid cells. As a result, the WRF-TEB upwelling longwave radiation flux density is overpredicted by about 100 W m⁻² (Figure 9e). The warmer WRF-UNN surface temperature can explain the larger turbulent sensible heat flux density at the observational site (Figure 9e) and across the domain (Figures 10e and 10f).

**Figure 9**. As Figure 8, but for online simulations.

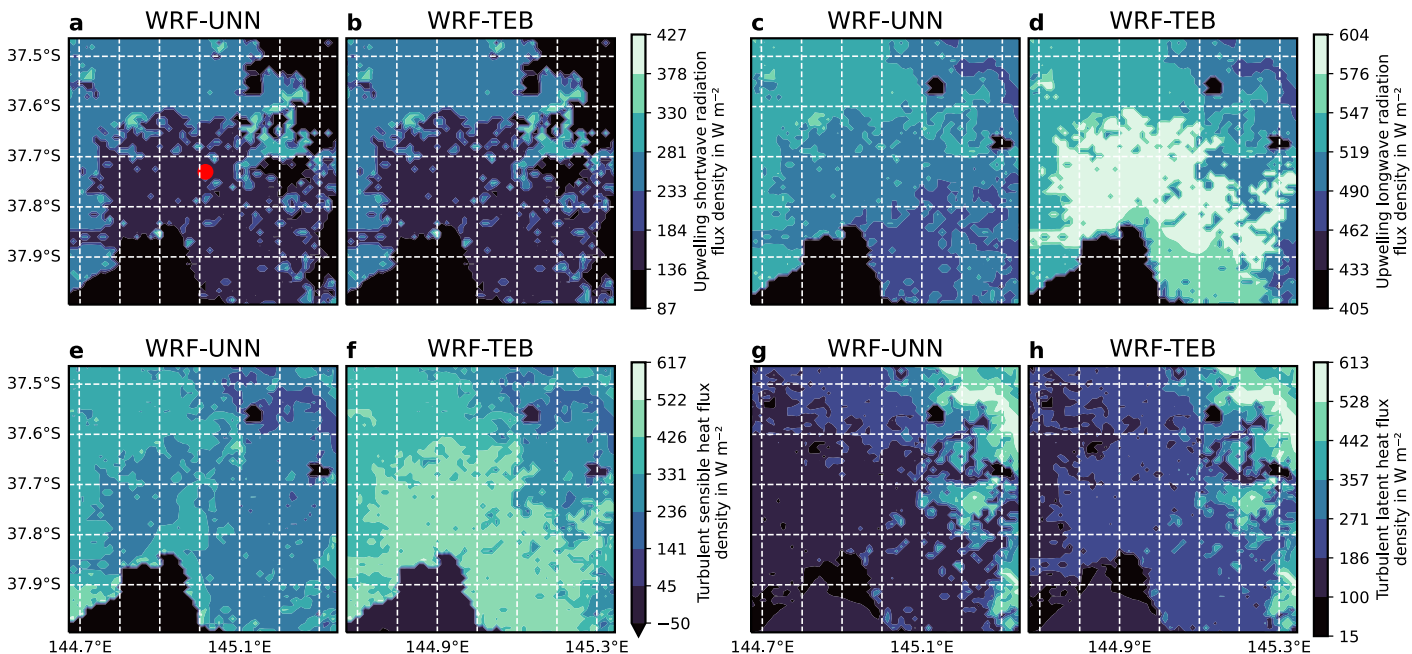

**Figure 10**. The inner domain (d4, Figure 5) with observation tower (red) for the 30-min average period between 12:00 and 12:30 on 25 December 2003 with flux densities simulated using (**a,c,e,g**) WRF-UNN and (**b,d,f,h**) WRF-TEB for (**a-b**) upwelling shortwave (**c-d**) and longwave radiation, and turbulent (**e-f**) sensible and (**g-h**) latent heat.

## 3.4 Computational Performance

The runtime between offline UNN and TEB simulations is compared based on 100 repeats each for the 16-month period. UNN runs include data normalization and inference using TensorFlow in Python. All runs are conducted on a shared AMD EPYC 7742 CPU node with 32 cores and 124 GiB of system memory. Both TEB and the UNN are configured to run fully single-threaded in a Singularity container running Ubuntu 20.04, GNU Fortran 9.3 compiler, and Anaconda Python 3.9. UNN ($0.50 \pm 0.0053$ s) runs are over one order of magnitude faster than TEB runs ($6.0 \pm 0.042$ s).

## 4. Conclusion

In this work, we successfully develop a neural network emulator of urban land surface processes (UNN) for offline and online applications. The UNN is trained on the multi-model ensemble mean (MEM) of 22 urban land surface models (ULSMs) for an area of Melbourne, Australia. The accuracy is assessed using flux observations and compared to a well-known ULSM (Town Energy Balance TEB) model. The MEM data are derived from a study with four Stages of increasing complexity (1–4; Appendix A). The UNN is trained using Stage 2 MEM, but compared to the Stage 4 TEB simulations, the latter using more site-specific information.

Compared to MEM, the UNN captures the general variability of surface energy balance fluxes. Relative to the observations, the UNN is more accurate than TEB—or than WRF-TEB when coupled to the Weather Research and Forecasting (WRF) model—while having reduced both computational demands (by over an order of magnitude) and model parameter requirements (i.e., trained using fewer site-specific parameters). Technically, the coupling to WRF is straightforward thanks to WRF-CMake and TensorFlow Lite C bindings.

As the first study to show the development and application of a machine learning (ML) emulator for urban land surface fluxes, we demonstrate its potential to improve the modeling of key surface energy balance fluxes: we combine the strengths of several ULSMs into one and show that such models can be successfully integrated into complex weather models, such as WRF. The development of (coupled) emulators such as WRF-UNN have other advantages compared to 'more-traditional' ULSMs such as code optimization at the deployment stage, and integration into different codebases and hardware architectures through common high-level APIs.

Although the current evaluation did not assess, or assume, surface energy balance closure, which is essential for climate applications (Grimmond et al., 2010), further research is needed to assess this before UNNs are used in climate studies. Furthermore, with no variations in urban areas (e.g., land cover fractions, surface parameters, and climate) assessed because of the current lack of multi-site data sets, the natural progression to assess our findings more globally requires data sets currently being developed (Lipson et al., 2020) with or without data augmentation strategies as outlined by Meyer et al. (2021).

Indeed, if MEMs are found to be more accurate than any individual ULSM on a global scale, an ML emulator as described here could help improve both the speed and accuracy of current ULSMs. Aside from the apparent speed-up improvement typical of ML emulators and that of improved accuracy outlined here, ML approaches may also prove helpful in operational Earth system models as the fewer site-specific parameters contained in MEM are easily retrievable and updatable globally using remote sensing techniques.

## Code and Data Availability

Software and tools are archived with a Singularity (Kurtzer et al., 2017) image and deposited on Zenodo as described in the scientific reproducibility section of Meyer, Schoetter, Riechert, et al. (2020). Users wishing to download (and reproduce) the results described in this paper can download the data archive at https://doi.org/10.5281/zenodo.5142960 (Meyer, 2021) and optionally run Singularity on their local or remote systems. Because of licensing restrictions, meteorological forcing (MF), and observational (OBS) and multi-model output (MO) data sets cannot be bundled with the Meyer (2021) data archive and need to be requested separately at https://doi.org/10.5281/zenodo.4679279 (Grimmond et al., 2021) and at https://doi.org/10.5281/zenodo.4678387 (Grimmond et al., 2013), respectively.

## Author Contributions

Conceptualization: D.M.; Data curation: D.M.; Formal analysis: D.M.; Investigation: D.M.; Methodology: D.M.; Software: D.M.; Resources: D.M.; Validation: D.M; Visualization: D.M.; Writing – original draft preparation: D.M.; Writing – review & editing: D.M., S.G., P.D., R.H., M.v.R..

## Acknowledgements

The authors would like to thank Stephen Rasp, Brian Henn (Allen Institute for Artificial Intelligence, Climate Modeling Group, Seattle, WA), and one anonymous reviewer for their valuable comments and feedback. The authors would also like to thank Andrew Coutts and Jason Beringer for supplying the observation data used in evaluations and those who contributed to the urban model comparison project. Peter Dueben gratefully acknowledges funding from the Royal Society for his University Research Fellowship, as well as from the ESiWACE Horizon 2020 project (#823988) and the MAELSTROM EuroHPC Joint Undertaking project (#955513).

## Appendix A: Stage Selection

Stage 2 data (Table S3) are selected for training the urban neural network (UNN; Section 2.2) and Stage 4 for the TEB to offer the best tradeoff between complexity and accuracy and better model metrics (Figure A1). Overall, TEB's mean bias and mean absolute error improve the more information is provided, notably when the site albedo is given in Stage 3 and 4 (Table S3). MEM generally has the lowest overall mean bias and mean absolute error for Stage 2.

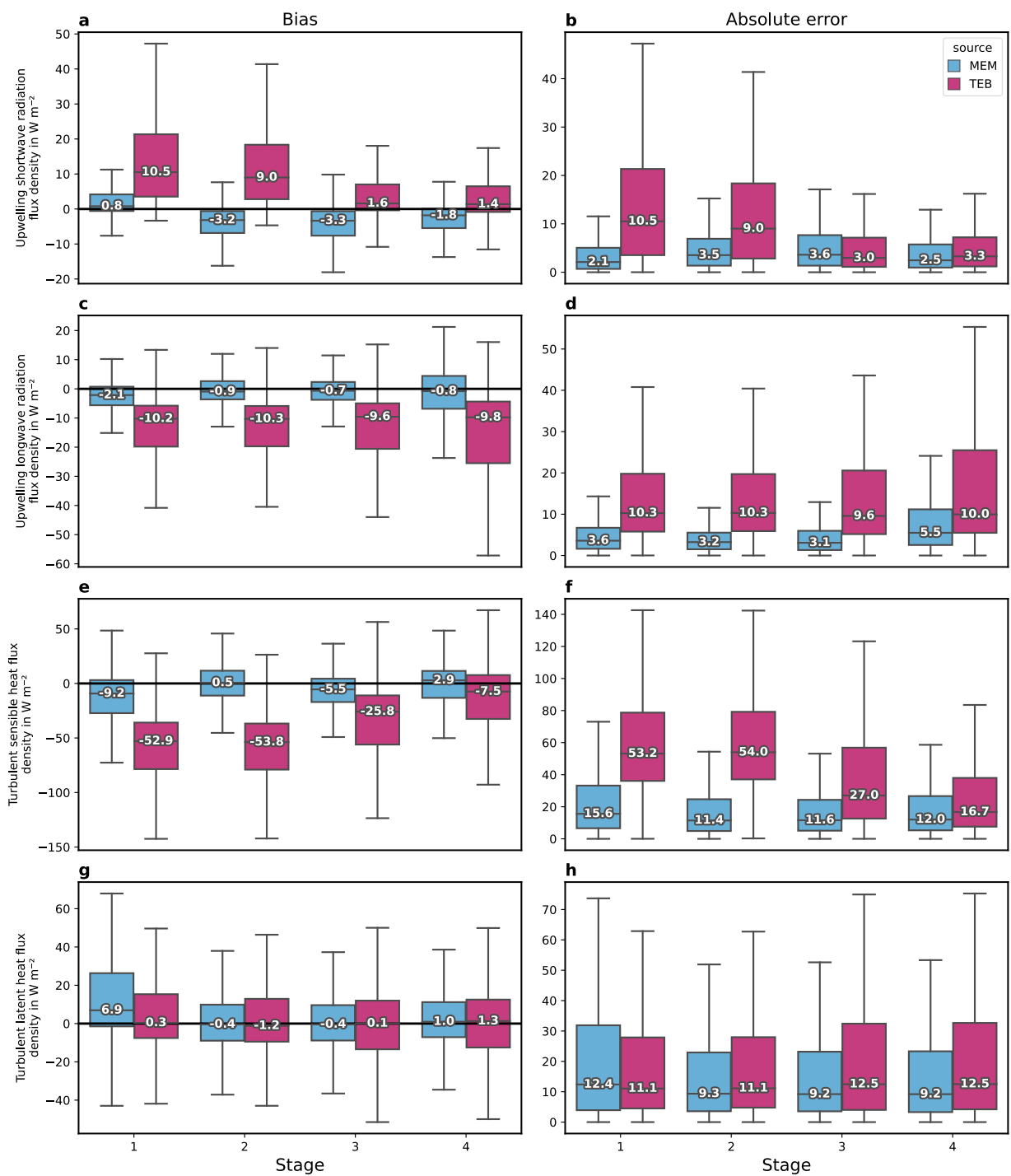

**Figure A1**. Distribution (boxplot with median (labels) interquartile range (IQR) and whiskers: 1.5 IQR) of the 30-min biases and absolute errors (section 2.6) relative to observations (OBS) for four Stages (Table S3) of multi-model ensemble mean (MEM) and Town Energy Balance (TEB) calculated for the test fraction (section 2.5) of 30-min fluxes of upwelling (**a-b**) short- and (**c-d**) long-wave radiation, and turbulent (**e-f**) sensible and (**g-h**) latent heat.

## References

Abadi, M., Agarwal, A., Barham, P., Brevdo, E., Chen, Z., Citro, C., Corrado, G. S., Davis, A., Dean, J., Devin, M., Ghemawat, S., Goodfellow, I., Harp, A., Irving, G., Isard, M., Jia, Y., Jozefowicz, R., Kaiser, L., Kudlur, M., … Zheng, X. (2015). *TensorFlow: Large-scale machine learning on heterogeneous systems*. https://www.tensorflow.org/

Baklanov, A., Grimmond, C. S. B., Carlson, D., Terblanche, D., Tang, X., Bouchet, V., Lee, B., Langendijk, G., Kolli, R. K., & Hovsepyan, A. (2018). From urban meteorology, climate and environment research to integrated city services. *Urban Climate*, *23*, 330–341. https://doi.org/10.1016/j.uclim.2017.05.004

Bengtsson, L., Andrae, U., Aspelien, T., Batrak, Y., Calvo, J., de Rooy, W., Gleeson, E., Hansen-Sass, B., Homleid, M., Hortal, M., Ivarsson, K.-I., Lenderink, G., Niemelä, S., Nielsen, K. P., Onvlee, J., Rontu, L., Samuelsson, P., Muñoz, D. S., Subias, A., … Køltzow, M. Ø. (2017). The HARMONIE–AROME Model Configuration in the ALADIN–HIRLAM NWP System. *Monthly Weather Review*, *145*(5), 1919–1935. https://doi.org/10.1175/MWR-D-16-0417.1

Best, M. J. (2005). Representing urban areas within operational numerical weather prediction models. *Boundary-Layer Meteorology*, *114*(1), 91–109. https://doi.org/10.1007/s10546-004-4834-5

Best, M. J., Grimmond, C. S. B., & Villani, M. G. (2006). Evaluation of the Urban Tile in MOSES using Surface Energy Balance Observations. *Boundary-Layer Meteorology*, *118*(3), 503–525. https://doi.org/10.1007/s10546-005-9025-5

Best, M. J., Pryor, M., Clark, D. B., Rooney, G. G., Essery, R . L. H., Ménard, C. B., Edwards, J. M., Hendry, M. A., Porson, A., Gedney, N., Mercado, L. M., Sitch, S., Blyth, E., Boucher, O., Cox, P. M., Grimmond, C. S. B., & Harding, R. J. (2011). The Joint UK Land Environment Simulator (JULES), model description – Part 1: Energy and water fluxes. *Geoscientific Model Development*, *4*(3), 677–699. https://doi.org/10.5194/gmd-4-677-2011

Bishop, C. M. (2006). *Pattern recognition and machine learning*. Springer.

Bolton, T., & Zanna, L. (2019). Applications of Deep Learning to Ocean Data Inference and Subgrid Parameterization. *Journal of Advances in Modeling Earth Systems*, *11*(1), 376–399. https://doi.org/10.1029/2018MS001472

Chen, F., & Dudhia, J. (2001). Coupling an Advanced Land Surface–Hydrology Model with the Penn State–NCAR MM5 Modeling System. Part I: Model Implementation and Sensitivity. *Monthly Weather Review*, *129*, 17. https://doi.org/10.1175/1520-0493(2001)129<0569:CAALSH>2.0.CO;2

Chen, F., Kusaka, H., Tewari, M., Bao, J., & Hirakuchi, H. (2004). Utilizing the coupled WRF/LSM/Urban modeling system with detailed urban classification to simulate the urban heat island phenomena over the Greater Houston area. *Fifth Symposium on the Urban Environment*, *25*, 9–11.

Coutts, A. M., Beringer, J., & Tapper, N. J. (2007a). Characteristics influencing the variability of urban CO2 fluxes in Melbourne, Australia. *Atmospheric Environment*, *41*(1), 51–62. https://doi.org/10.1016/j.atmosenv.2006.08.030

Coutts, A. M., Beringer, J., & Tapper, N. J. (2007b). Impact of Increasing Urban Density on Local Climate: Spatial and Temporal Variations in the Surface Energy Balance in Melbourne, Australia. *Journal of Applied Meteorology and Climatology*, *46*(4), 477–493. https://doi.org/10.1175/JAM2462.1

Dandou, A. (2005). Development and evaluation of an urban parameterization scheme in the Penn State/NCAR Mesoscale Model (MM5). *Journal of Geophysical Research*, *110*(D10), D10102. https://doi.org/10.1029/2004JD005192

Dupont, S., & Mestayer, P. G. (2006). Parameterization of the Urban Energy Budget with the Submesoscale Soil Model. *Journal of Applied Meteorology and Climatology*, *45*(12), 1744–1765. https://doi.org/10.1175/JAM2417.1

Dupont, S., Mestayer, P. G., Guilloteau, E., Berthier, E., & Andrieu, H. (2006). Parameterization of the Urban Water Budget with the Submesoscale Soil Model. *Journal of Applied Meteorology and Climatology*, *45*(4), 624–648. https://doi.org/10.1175/JAM2363.1

ECMWF. (2004). *Cycle 28r2 summary of changes* [Text]. ECMWF Documentation. https://www.ecmwf.int/en/forecasts/documentation-and-support/evolution-ifs/cycle-archived/cycle-28r2-summary-changes

Essery, R. L. H., Best, M. J., Betts, R. A., Cox, P. M., & Taylor, C. M. (2003). Explicit Representation of Subgrid Heterogeneity in a GCM Land Surface Scheme. *Journal of Hydrometeorology*, *4*(3), 530–543. https://doi.org/10.1175/1525-7541(2003)004<0530:EROSHI>2.0.CO;2

Fortuniak, K. (2003). A slab surface energy balance model (SUEB) and its application to the study on the role of roughness length in forming an urban heat island. *Acta Universitatis Wratislaviensis*, *2542*, 368–377.

Fortuniak, K., Offerle, B., & Grimmond, C. (2005). Application of a slab surface energy balance model to determine surface parameters for urban areas. *Lund Electronic Reports in Physical Geography*, *5*, 90–91.

Goodfellow, I., Bengio, Y., & Courville, A. (2016). *Deep learning*. MIT Press.

Grimmond, C. S. B., Best, M., Barlow, J., Arnfield, A. J., Baik, J.-J., Baklanov, A., Belcher, S., Bruse, M., Calmet, I., Chen, F., Clark, P., Dandou, A., Erell, E., Fortuniak, K., Hamdi, R., Kanda, M., Kawai, T., Kondo, H., Krayenhoff, S., … Williamson, T. (2009). Urban Surface Energy Balance Models: Model Characteristics and Methodology for a Comparison Study. In A.

Baklanov, G. Sue, M. Alexander, & M. Athanassiadou (Eds.), *Meteorological and Air Quality Models for Urban Areas* (pp. 97–123). Springer Berlin Heidelberg. https://doi.org/10.1007/978-3-642-00298-4_11

Grimmond, C. S. B., Blackett, M., Best, M. J., Baik, J.-J., Belcher, S. E., Beringer, J., Bohnenstengel, S. I., Calmet, I., Chen, F., Coutts, A., Dandou, A., Fortuniak, K., Gouvea, M. L., Hamdi, R., Hendry, M., Kanda, M., Kawai, T., Kawamoto, Y., Kondo, H., … Zhang, N. (2011). Initial results from Phase 2 of the international urban energy balance model comparison. *International Journal of Climatology*, *31*(2), 244–272. https://doi.org/10.1002/joc.2227

Grimmond, C. S. B., Blackett, M., Best, M. J., Barlow, J., Baik, J.-J., Belcher, S. E., Bohnenstengel, S. I., Calmet, I., Chen, F., Dandou, A., Fortuniak, K., Gouvea, M. L., Hamdi, R., Hendry, M., Kawai, T., Kawamoto, Y., Kondo, H., Krayenhoff, E. S., Lee, S.-H., … Zhang, N. (2010). The International Urban Energy Balance Models Comparison Project: First Results from Phase 1. *Journal of Applied Meteorology and Climatology*, *49*(6), 1268–1292. https://doi.org/10.1175/2010JAMC2354.1

Grimmond, C. S. B., & Oke, T. R. (2002). Turbulent Heat Fluxes in Urban Areas: Observations and a Local-Scale Urban Meteorological Parameterization Scheme (LUMPS). *Journal of Applied Meteorology*, *41*, 19. https://doi.org/10.1175/1520-0450(2002)041<0792:THFIUA>2.0.CO;2

Grimmond, S., Blackett, M., Best, M., Coutts, A., Beringer, J., & Urban Model Comparison Team. (2021). *Phase 2 of the International urban energy balance comparison project—Forcing Data* (Original) [Data set]. Zenodo. https://doi.org/10.5281/zenodo.4679279

Grimmond, S., Blackett, M., Best, M., & Urban Model Comparison Team. (2013). *Phase 2 of the International urban energy balance comparison project—Data analysed—Anonymous* (Anonymous (10/4/21)) [Data set]. Zenodo. https://doi.org/10.5281/zenodo.4678387

Grimmond, S., Bouchet, V., Molina, L. T., Baklanov, A., Tan, J., Schlünzen, K. H., Mills, G., Golding, B., Masson, V., Ren, C., Voogt, J., Miao, S., Lean, H., Heusinkveld, B., Hovespyan, A., Teruggi, G., Parrish, P., & Joe, P. (2020). Integrated urban hydrometeorological, climate and environmental services: Concept, methodology and key messages. *Urban Climate*, *33*, 100623. https://doi.org/10.1016/j.uclim.2020.100623

Hamdi, R., Degrauwe, D., & Termonia, P. (2012). Coupling the Town Energy Balance (TEB) Scheme to an Operational Limited-Area NWP Model: Evaluation for a Highly Urbanized Area in Belgium. *Weather and Forecasting*, *27*(2), 323–344. https://doi.org/10.1175/WAF-D-11-00064.1

Hamdi, R., & Masson, V. (2008). Inclusion of a Drag Approach in the Town Energy Balance (TEB) Scheme: Offline 1D Evaluation in a Street Canyon. *Journal of Applied Meteorology and Climatology*, *47*(10), 2627–2644. https://doi.org/10.1175/2008JAMC1865.1

Harman, I. N., Barlow, J. F., & Belcher, S. E. (2004). Scalar Fluxes from Urban Street Canyons Part II: Model. *Boundary-Layer Meteorology*, *113*(3), 387–410. https://doi.org/10.1007/s10546-004-6205-7

Harman, I. N., & Belcher, S. E. (2006). The surface energy balance and boundary layer over urban street canyons. *Quarterly Journal of the Royal Meteorological Society*, *132*(621), 2749–2768. https://doi.org/10.1256/qj.05.185

Harman, I. N., Best, M. J., & Belcher, S. E. (2004). Radiative Exchange in an Urban Street Canyon. *Boundary-Layer Meteorology*, *110*(2), 301–316. https://doi.org/10.1023/A:1026029822517

Hertwig, D., Ng, M., Grimmond, S., Vidale, P. L., & McGuire, P. C. (2021). High-resolution global climate simulations: Representation of cities. *International Journal of Climatology*, *41*(5), 3266–3285. https://doi.org/10.1002/joc.7018

Holmgren, W. F., Calama-Consulting, Hansen, C., Anderson, K., Mikofski, M., Lorenzo, A., Krien, U., Bmu, Stark, C., DaCoEx, Driesse, A., Jensen, A. R., De León Peque, M. S., Konstant_t, Mayudong, Heliolytics, Miller, E., Anoma, M. A., Guo, V., … Dollinger, J. (2021). *pvlib/pvlib-python: V0.9.0* (v0.9.0) [Computer software]. Zenodo. https://doi.org/10.5281/ZENODO.5366883

Holmgren, W. F., Hansen, C. W., & Mikofski, M. A. (2018). pvlib python: A python package for modeling solar energy systems. *Journal of Open Source Software*, *3*(29), 884. https://doi.org/10.21105/joss.00884

Hong, S.-Y., & Lim, J.-O. J. (2006). The WRF Single-Moment 6-Class Microphysics Scheme (WSM6). *Journal of the Korean Meteorological Society*, *42*(2), 129–151.

Hong, S.-Y., Noh, Y., & Dudhia, J. (2006). A New Vertical Diffusion Package with an Explicit Treatment of Entrainment Processes. *Monthly Weather Review*, *134*(9), 2318–2341. https://doi.org/10.1175/MWR3199.1

Iacono, M. J., Delamere, J. S., Mlawer, E. J., Shephard, M. W., Clough, S. A., & Collins, W. D. (2008). Radiative forcing by long-lived greenhouse gases: Calculations with the AER radiative transfer models. *Journal of Geophysical Research*, *113*(D13), D13103. https://doi.org/10.1029/2008JD009944

Jiménez, P. A., Dudhia, J., González-Rouco, J. F., Navarro, J., Montávez, J. P., & García-Bustamante, E. (2012). A Revised Scheme for the WRF Surface Layer Formulation. *Monthly Weather Review*, *140*(3), 898–918. https://doi.org/10.1175/MWR-D-11-00056.1

Kanda, M., Kawai, T., Kanega, M., Moriwaki, R., Narita, K., & Hagishima, A. (2005). A Simple Energy Balance Model for Regular Building Arrays. *Boundary-Layer Meteorology*, *116*(3), 423–443. https://doi.org/10.1007/s10546-004-7956-x

Kanda, M., Kawai, T., & Nakagawa, K. (2005). A Simple Theoretical Radiation Scheme for Regular Building Arrays. *Boundary-Layer Meteorology*, *114*(1), 71–90. https://doi.org/10.1007/s10546-004-8662-4

Kawai, T., Kanda, M., Narita, K., & Hagishima, A. (2007). Validation of a numerical model for urban energy-exchange using outdoor scale-model measurements. *International Journal of Climatology*, *27*(14), 1931–1942. https://doi.org/10.1002/joc.1624

Kawai, T., Ridwan, M. K., & Kanda, M. (2009). Evaluation of the Simple Urban Energy Balance Model Using Selected Data from 1-yr Flux Observations at Two Cities. *Journal of Applied Meteorology and Climatology*, *48*(4), 693–715. https://doi.org/10.1175/2008JAMC1891.1

Kawamoto, Y., & Ooka, R. (2006). Analysis of the radiation field at pedestrian level using a meso-scale meteorological model incorporating the urban canopy model. *Sixth International Conference on Urban Climate*, 446–449.

Kawamoto, Y., & Ooka, R. (2009a). Accuracy validation of urban climate analysis model using MM5 incorporating a multi-layer urban canopy model. *Seventh International Conference on Urban Climate*, 4.

Kawamoto, Y., & Ooka, R. (2009b). Incorporating an urban canopy model to represent the effects of buildings: Development of urban climate analysis model using MM5 Part 2. *Journal of Environmental Engineering (Transactions of AIJ)*, *74*(642), 1009–1018. https://doi.org/10.3130/aije.74.1009

Kelso, N. V., & Patterson, T. (2009). Natural Earth Vector. *Cartographic Perspectives*, *64*, 45–50. https://doi.org/10.14714/CP64.148

Kingma, D. P., & Ba, J. (2015). *Adam: A Method for Stochastic Optimization*. 3rd International Conference on Learning Representations (ICLR), San Diego, CA, USA. https://arxiv.org/abs/1412.6980

Kondo, H., Genchi, Y., Kikegawa, Y., Ohashi, Y., Yoshikado, H., & Komiyama, H. (2005). Development of a Multi-Layer Urban Canopy Model for the Analysis of Energy Consumption in a Big City: Structure of the Urban Canopy Model and its Basic Performance. *Boundary-Layer Meteorology*, *116*(3), 395–421. https://doi.org/10.1007/s10546-005-0905-5

Kondo, H., & Liu, F.-H. (1998). A study on the urban thermal environment obtained through one-dimensional urban canopy model. *Journal of Japan Society for Atmospheric Environment*, *33*(3), 179–192. https://doi.org/10.11298/taiki1995.33.3_179

Krasnopolsky, V. M., Fox-Rabinovitz, M. S., & Belochitski, A. A. (2013). Using Ensemble of Neural Networks to Learn Stochastic Convection Parameterizations for Climate and Numerical Weather Prediction Models from Data Simulated by a Cloud Resolving Model. *Advances in Artificial Neural Systems*, *2013*, 1–13. https://doi.org/10.1155/2013/485913

Krayenhoff, E. S., & Voogt, J. A. (2007). A microscale three-dimensional urban energy balance model for studying surface temperatures. *Boundary-Layer Meteorology*, *123*(3), 433–461. https://doi.org/10.1007/s10546-006-9153-6

Kurtzer, G. M., Sochat, V., & Bauer, M. W. (2017). Singularity: Scientific containers for mobility of compute. *PLOS ONE*, *12*(5), e0177459. https://doi.org/10.1371/journal.pone.0177459

Kusaka, H., Kondo, H., Kikegawa, Y., & Kimura, F. (2001). A Simple Single-Layer Urban Canopy Model For Atmospheric Models: Comparison With Multi-Layer And Slab Models. *Boundary-Layer Meteorology*, *101*(3), 329–358. https://doi.org/10.1023/A:1019207923078

Lee, S.-H., & Park, S.-U. (2007). A Vegetated Urban Canopy Model for Meteorological and Environmental Modelling. *Boundary-Layer Meteorology*, *126*(1), 73–102. https://doi.org/10.1007/s10546-007-9221-6

Lemonsu, A., Grimmond, C. S. B., & Masson, V. (2004). Modeling the Surface Energy Balance of the Core of an Old Mediterranean City: Marseille. *Journal of Applied Meteorology*, *43*, 16. https://doi.org/10.1175/1520-0450(2004)043<0312:MTSEBO>2.0.CO;2

Lemonsu, A., & Masson, V. (2002). Simulation of a Summer Urban Breeze Over Paris. *Boundary-Layer Meteorology*, *104*(3), 463–490. https://doi.org/10.1023/A:1016509614936

Li, L., Jamieson, K., DeSalvo, G., Rostamizadeh, A., & Talwalkar, A. (2018). Hyperband: A novel bandit-based approach to hyperparameter optimization. *Journal of Machine Learning Research*, *18*(185), 1–52. http://jmlr.org/papers/v18/16-558.html

Lipson, M., Grimmond, S., & Best, M. (2020). Calling for participants: A new multi-site evaluation project for modelling in urban areas. *Urban Climate News: Quarterly Newsletter of the IAUC*, *75*, 15–16. https://www.urban-climate.org/wp-content/uploads/IAUC075.pdf

Loridan, T., Grimmond, C. S. B., Grossman-Clarke, S., Chen, F., Tewari, M., Manning, K., Martilli, A., Kusaka, H., & Best, M. (2010). Trade-offs and responsiveness of the single-layer urban canopy parametrization in WRF: An offline evaluation using the MOSCEM optimization algorithm and field observations. *Quarterly Journal of the Royal Meteorological Society*, *136*(649), 997–1019. https://doi.org/10.1002/qj.614

Loridan, T., Grimmond, C. S. B., Offerle, B. D., Young, D. T., Smith, T. E. L., Järvi, L., & Lindberg, F. (2011). Local-Scale Urban Meteorological Parameterization Scheme (LUMPS): Longwave Radiation Parameterization and Seasonality-Related

Developments. *Journal of Applied Meteorology and Climatology*, *50*(1), 185–202. https://doi.org/10.1175/2010JAMC2474.1

Martilli, A., Clappier, A., & Rotach, M. W. (2002). An Urban Surface Exchange Parameterisation for Mesoscale Models. *Boundary-Layer Meteorology*, *104*(2), 261–304. https://doi.org/10.1023/A:1016099921195

Masson, V. (2000). A Physically-Based Scheme For The Urban Energy Budget In Atmospheric Models. *Boundary-Layer Meteorology*, *94*(3), 357–397. https://doi.org/10.1023/A:1002463829265

Masson, V., Grimmond, C. S. B., & Oke, T. R. (2002). Evaluation of the Town Energy Balance (TEB) Scheme with Direct Measurements from Dry Districts in Two Cities. *Journal of Applied Meteorology*, *41*, 16. https://doi.org/10.1175/1520-0450(2002)041<1011:EOTTEB>2.0.CO;2

Masson, V., Lemonsu, A., Pigeon, G., Schoetter, R., de Munck, C., Bueno, B., Faroux, S., Goret, M., Redon, E., Chancibault, K., Stavropulos-Laffaille, X., Leroyer, S., & Meyer, D. (2021). *The Town Energy Balance (TEB) model* (4.1.2) [Computer software]. Zenodo. https://doi.org/10.5281/zenodo.5775962

May, R. M., Arms, S. C., Marsh, P., Bruning, E., Leeman, J. R., Goebbert, K., Thielen, J. E., Bruick, Z. S., & Camron, M. Drew. (2021). *MetPy: A Python package for meteorological data* (1.1.0) [Computer software]. https://doi.org/10.5065/D6WW7G29

Meyer, D. (2021). *Data archive for paper "Machine Learning Emulation of Urban Land Surface Processes"* (1.0.0) [Data set]. Zenodo. https://doi.org/10.5281/zenodo.5142960

Meyer, D., Hogan, R. J., Dueben, P. D., & Mason, S. L. (2022). Machine Learning Emulation of 3D Cloud Radiative Effects. *Journal of Advances in Modeling Earth Systems*, *14*(3). https://doi.org/10.1029/2021MS002550

Meyer, D., Nagler, T., & Hogan, R. J. (2021). Copula-based synthetic data augmentation for machine-learning emulators. *Geoscientific Model Development*, *14*(8), 5205–5215. https://doi.org/10.5194/gmd-14-5205-2021

Meyer, D., & Riechert, M. (2019). Open source QGIS toolkit for the Advanced Research WRF modelling system. *Environmental Modelling & Software*, *112*, 166–178. https://doi.org/10.1016/j.envsoft.2018.10.018

Meyer, D., & Riechert, M. (2020). *The GIS4WRF Plugin* (0.14.4) [Computer software]. Zenodo. https://doi.org/10.5281/zenodo.4035877

Meyer, D., Schoetter, R., Masson, V., & Grimmond, S. (2020). Enhanced software and platform for the Town Energy Balance (TEB) model. *Journal of Open Source Software*, *5*(50), 2008. https://doi.org/10.21105/joss.02008

Meyer, D., Schoetter, R., Riechert, M., Verrelle, A., Tewari, M., Dudhia, J., Masson, V., van Reeuwijk, M., & Grimmond, S. (2020). WRF-TEB: Implementation and Evaluation of the Coupled Weather Research and Forecasting (WRF) and Town Energy Balance (TEB) Model. *Journal of Advances in Modeling Earth Systems*, *12*(8). https://doi.org/10.1029/2019MS001961

Meyer, D., & Thevenard, D. (2019). PsychroLib: A library of psychrometric functions to calculate thermodynamic properties of air. *Journal of Open Source Software*, *4*(33), 1137. https://doi.org/10.21105/joss.01137

Meyer, D., & Thevenard, D. (2020). *PsychroLib: A library of psychrometric functions to calculate thermodynamic properties of air* (2.5.0) [Computer software]. Zenodo. https://doi.org/10.5281/zenodo.3748874

Nowack, P., Braesicke, P., Haigh, J., Abraham, N. L., Pyle, J., & Voulgarakis, A. (2018). Using machine learning to build temperature-based ozone parameterizations for climate sensitivity simulations. *Environmental Research Letters*, *13*(10), 104016. https://doi.org/10.1088/1748-9326/aae2be

Offerle, B., Grimmond, C. S. B., & Oke, T. R. (2003). Parameterization of Net All-Wave Radiation for Urban Areas. *Journal of Applied Meteorology*, *42*, 17. https://doi.org/10.1175/1520-0450(2003)042<1157:PONARF>2.0.CO;2

Oke, T. R., Mills, G., Christen, A., & Voogt, J. A. (2017). *Urban Climates*. Cambridge University Press. https://doi.org/10.1017/9781139016476

Oleson, K. W., Bonan, G. B., Feddema, J., & Jackson, T. (2011). An examination of urban heat island characteristics in a global climate model: HEAT ISLAND CHARACTERISTICS IN A GLOBAL CLIMATE MODEL. *International Journal of Climatology*, *31*(12), 1848–1865. https://doi.org/10.1002/joc.2201

Oleson, K. W., Bonan, G. B., Feddema, J., & Vertenstein, M. (2008). An Urban Parameterization for a Global Climate Model. Part II: Sensitivity to Input Parameters and the Simulated Urban Heat Island in Offline Simulations. *Journal of Applied Meteorology and Climatology*, *47*(4), 1061–1076. https://doi.org/10.1175/2007JAMC1598.1

Oleson, K. W., Bonan, G. B., Feddema, J., Vertenstein, M., & Grimmond, C. S. B. (2008). An Urban Parameterization for a Global Climate Model. Part I: Formulation and Evaluation for Two Cities. *Journal of Applied Meteorology and Climatology*, *47*(4), 1038–1060. https://doi.org/10.1175/2007JAMC1597.1

O'Malley, T., Bursztein, E., Long, J., Chollet, F., Jin, H., Invernizzi, L., & others. (2019). *KerasTuner*. https://github.com/keras-team/keras-tuner

OpenStreetMap contributors. (2017). *Planet dump retrieved from https://planet.osm.org*. https://www.openstreetmap.org

Pigeon, G., Moscicki, M. A., Voogt, J. A., & Masson, V. (2008). Simulation of fall and winter surface energy balance over a dense urban area using the TEB scheme. *Meteorology and Atmospheric Physics*, *102*(3–4), 159–171. https://doi.org/10.1007/s00703-008-0320-9

Porson, A., Clark, P. A., Harman, I. N., Best, M. J., & Belcher, S. E. (2010). Implementation of a new urban energy budget scheme into MetUM. Part II: Validation against observations and model intercomparison. *Quarterly Journal of the Royal Meteorological Society*, *136*(651), 1530–1542. https://doi.org/10.1002/qj.572

Rasp, S., & Lerch, S. (2018). Neural Networks for Postprocessing Ensemble Weather Forecasts. *Monthly Weather Review*, *146*(11), 3885–3900. https://doi.org/10.1175/MWR-D-18-0187.1

Rasp, S., Pritchard, M. S., & Gentine, P. (2018). Deep learning to represent subgrid processes in climate models. *Proceedings of the National Academy of Sciences*, *115*(39), 9684–9689. https://doi.org/10.1073/pnas.1810286115

Riechert, M., & Meyer, D. (2019a). *WPS-CMake (Version 4.1.0)* (WPS-CMake-4.1.0) [Computer software]. Zenodo. https://doi.org/10.5281/zenodo.3407075

Riechert, M., & Meyer, D. (2019b). WRF-CMake: Integrating CMake support into the Advanced Research WRF (ARW) modelling system. *Journal of Open Source Software*, *4*(41), 1468. https://doi.org/10.21105/joss.01468

Riechert, M., & Meyer, D. (2021). *WRF-CMake: Integrating CMake support into the Advanced Research WRF (ARW) modelling system* (WRF-CMake-4.2.2) [Computer software]. Zenodo. https://doi.org/10.5281/zenodo.4449077

Ryu, Y.-H., Baik, J.-J., & Lee, S.-H. (2011). A New Single-Layer Urban Canopy Model for Use in Mesoscale Atmospheric Models. *Journal of Applied Meteorology and Climatology*, *50*(9), 1773–1794. https://doi.org/10.1175/2011JAMC2665.1

Salamanca, F., Krayenhoff, E. S., & Martilli, A. (2009). On the Derivation of Material Thermal Properties Representative of Heterogeneous Urban Neighborhoods. *Journal of Applied Meteorology and Climatology*, *48*(8), 1725–1732. https://doi.org/10.1175/2009JAMC2176.1

Salamanca, F., Krpo, A., Martilli, A., & Clappier, A. (2010). A new building energy model coupled with an urban canopy parameterization for urban climate simulations—Part I. formulation, verification, and sensitivity analysis of the model. *Theoretical and Applied Climatology*, *99*(3–4), 331–344. https://doi.org/10.1007/s00704-009-0142-9

Salamanca, F., & Martilli, A. (2010). A new Building Energy Model coupled with an Urban Canopy Parameterization for urban climate simulations—Part II. Validation with one dimension off-line simulations. *Theoretical and Applied Climatology*, *99*(3–4), 345–356. https://doi.org/10.1007/s00704-009-0143-8

Seity, Y., Brousseau, P., Malardel, S., Hello, G., Bénard, P., Bouttier, F., Lac, C., & Masson, V. (2011). The AROME-France Convective-Scale Operational Model. *Monthly Weather Review*, *139*(3), 976–991. https://doi.org/10.1175/2010MWR3425.1

Skamarock, W. C., Klemp, J. B., Dudhia, J., Gill, D. O., Liu, Z., Berner, J., Wang, W., Powers, J. G., Duda, M. G., Barker, D. M., & Huang, X.-Y. (2019). A Description of the Advanced Research WRF Model Version 4. *NCAR Technical Note NCAR/TN-556+STR*, 145. https://doi.org/10.5065/1dfh-6p97

Stamen Design. (2021). *Stamen Maps*. http://maps.stamen.com/

Stensrud, D. J. (2007). *Parameterization Schemes: Keys to Understanding Numerical Weather Prediction Models*. Cambridge University Press. https://doi.org/10.1017/CBO9780511812590

TensorFlow Developers. (2021a). *Install TensorFlow for C*. TensorFlow. https://www.tensorflow.org/install/lang_c

TensorFlow Developers. (2021b). *TensorFlow Lite | ML for Mobile and Edge Devices*. TensorFlow. https://www.tensorflow.org/lite

TensorFlow Developers. (2021c). *TensorFlow* (v2.6.2) [Computer software]. Zenodo. https://doi.org/10.5281/zenodo.5645375

UCAR. (2019). *WPS V4 Geographical Static Data*. http://www2.mmm.ucar.edu/wrf/users/download/get_sources_wps_geog.html

Zhang, C., & Wang, Y. (2017). Projected Future Changes of Tropical Cyclone Activity over the Western North and South Pacific in a 20-km-Mesh Regional Climate Model. *Journal of Climate*, *30*(15), 5923–5941. https://doi.org/10.1175/JCLI-D-16-0597.1

## Supplementary information (SI)

Original 32 (Figure S2) and the selected models (Figure S3) urban land surface model outputs with the Grimmond et al. (2011) anonymous identifiers for the four Stages (Table S3).

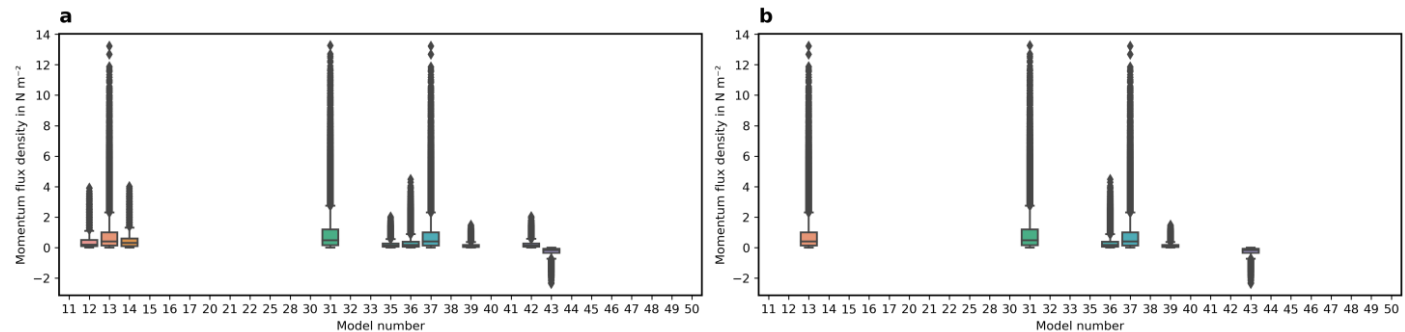

**Figure S1**. Momentum flux density for the (**a**) original and (**b**) reduced number of models outputs data set. As the original data set does not include the observations and only for six model, sit is not used in this study.

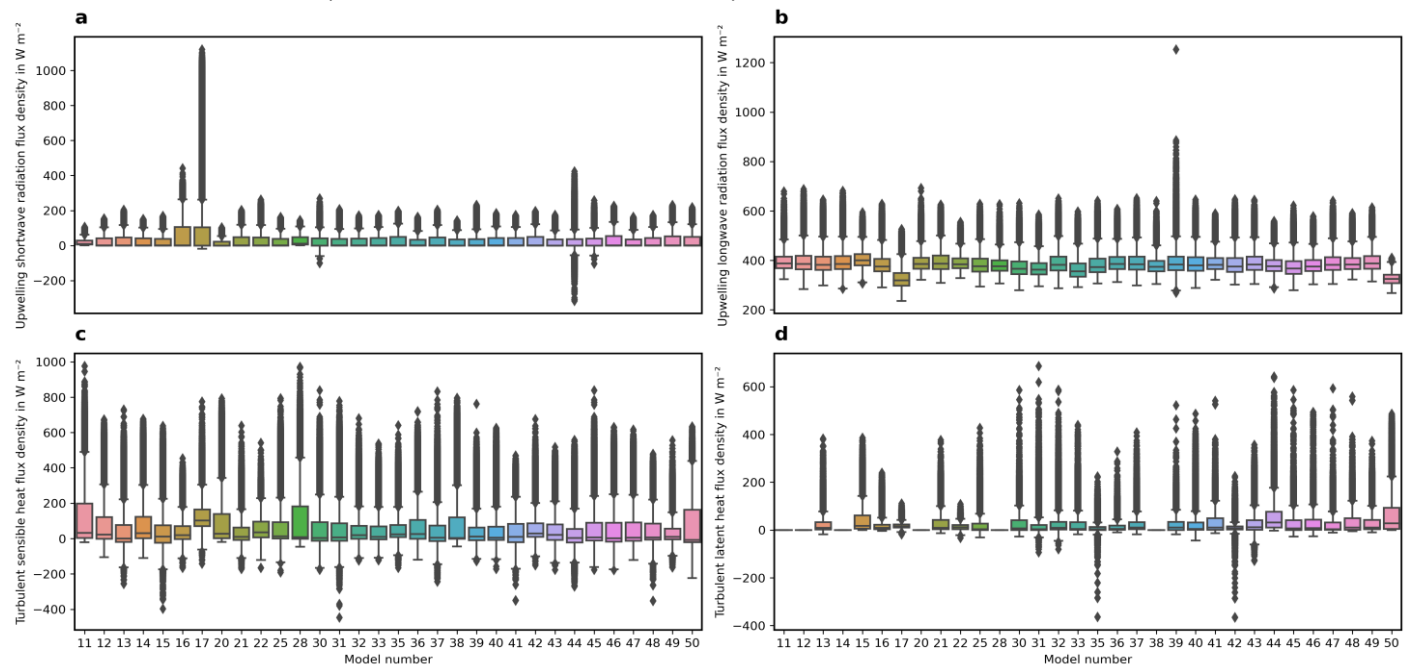

**Figure S2**. Fluxes modelled by 32 urban land surface models or configurations for all four Stages (Grimmond et al. 2011) for four fluxes: upwelling (**a**) short- and (**b**) long-wave radiation, turbulent (**c**) sensible and (**d**) latent heat.

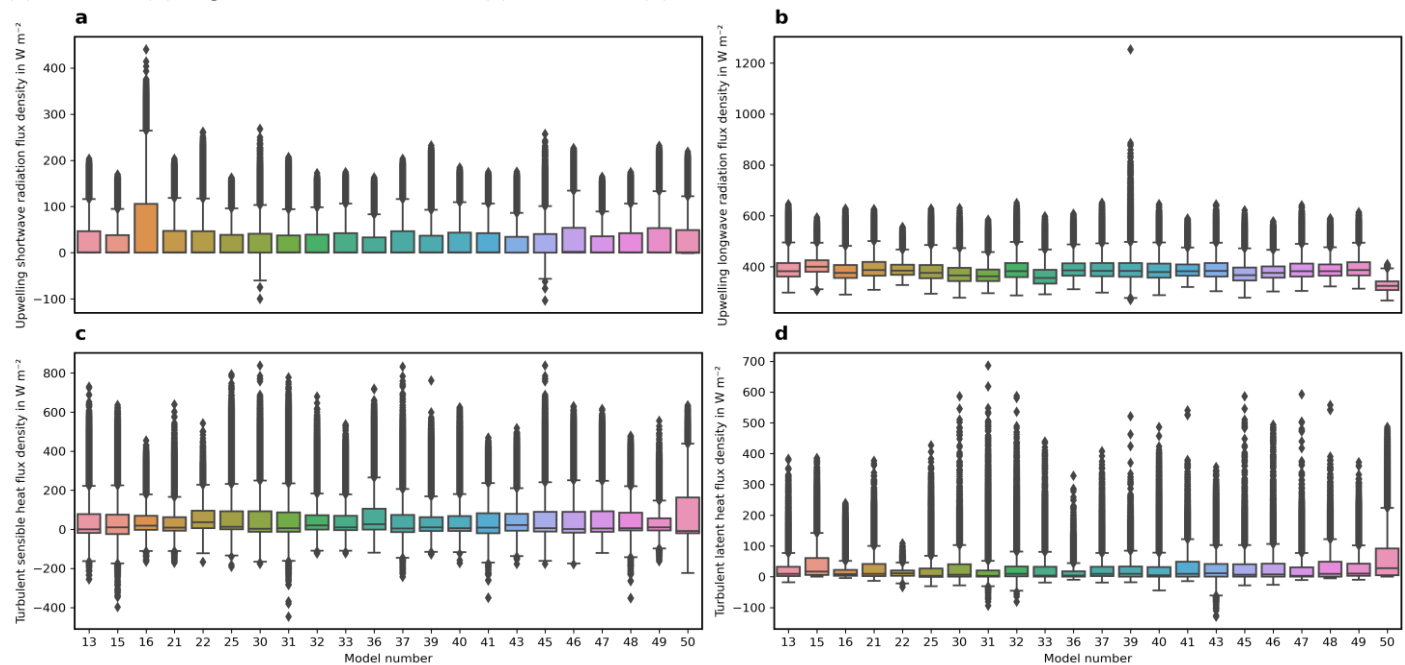

**Figure S3**. As Fig S2, but for the selected 22 urban land surface models. Poor performers (17 and 44), or models/configurations not including latent heat flux (11, 12, 14, 20, 28, 35, 38, 42) are removed.

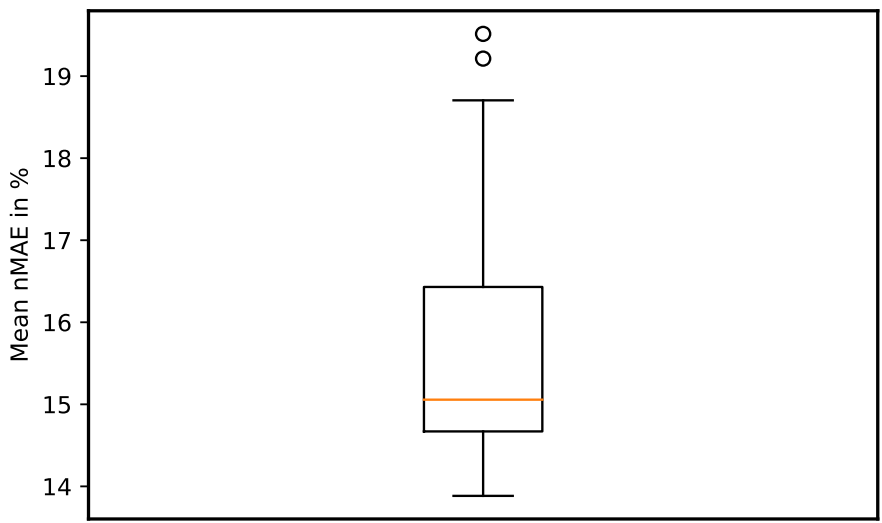

**Figure S4**. Boxplot of mean nMAE UNN iterations (section 2.5).

**Table S1.** Configuration used for the grid search in KerasTuner (O'Malley et al., 2019) version 1.0.4 using Hyperband (Li et al., 2018). The Adam algorithm (Kingma & Ba, 2015) with mean squared error as its optimization function is used. The epoch limit is set to 200 and the early stopping patience to 20 epochs using 25 % of the training fraction as defined in section 2.5.

| Hyperparameter | Values |
|---|---|
| Number of hidden layers | [1, 2, 3] |
| Number of Neurons | [16—512, every 32] |
| Activation function | [relu, tanh, sigmoid] |
| L2 Regularization | [$1e^{-3}$, $1e^{-2}$, $1e^{-1}$, $1e^{0}$, $1e^{1}$, $1e^{2}$] |

**Table S2**. Model participants in the Grimmond et al. (2011) study.

| Model | Reference |
|---|---|
| Building effect parameterization (BEP) | Martilli et al. (2002) |
| BEP coupled with building energy model | Martilli et al. (2002); Salamanca et al. (2009, 2010); Salamanca & Martilli (2010) |
| Community land model – urban (CLM-urban) | Oleson, Bonan, Feddema, & Vertenstein (2008); Oleson, Bonan, Feddema, Vertenstein, et al. (2008) |
| Institute of Industrial Science urban canopy model | Kawamoto & Ooka (2006, 2009a, 2009b) |
| Joint UK land environment simulator (JULES) | Best (2005); Best et al. (2006, 2011); Essery et al. (2003) |
| Local-scale urban meteorological parameterization scheme (LUMPS) | Grimmond & Oke (2002); Loridan et al.(2011); Offerle et al. (2003) |
| Met Office Reading urban surface exchange scheme (MORUSES) | Harman, Barlow, et al. (2004); Harman, Best, et al. (2004); Porson et al. (2010) |
| Multi-layer urban canopy model | Kondo et al. (2005); Kondo & Liu (1998) |
| Nanjing University urban canopy model-single layer | Kusaka et al. (2001); Masson (2000) |
| National and Kapodistrian University of Athens model | Dandou (2005) |
| Noah land surface model/single-layer urban canopy model | Chen et al. (2004); Kusaka et al. (2001); Loridan et al. (2010) |
| Seoul National University urban canopy model | Ryu et al. (2011) |
| Simple urban energy balance model for mesoscale simulation | Kanda, Kawai, Kanega, et al. (2005); Kanda, Kawai, & Nakagawa (2005); Kawai et al. (2007, 2009) |
| Single column Reading urban model tile version | Harman & Belcher (2006) |
| Slab urban energy balance model | Fortuniak (2003); Fortuniak et al. (2005) |
| Soil model for sub-meso scales (urbanised) | Dupont et al. (2006); Dupont & Mestayer (2006) |
| Temperatures of urban facets (TUF) 2D | Krayenhoff & Voogt (2007) |
| Temperatures of urban facets (TUF) 3D | Krayenhoff & Voogt (2007) |
| Town energy balance (TEB) | Lemonsu et al. (2004); Masson (2000); Masson et al. (2002); Pigeon et al. (2008) |
| Town energy balance (TEB) with multi-layer option | Hamdi & Masson (2008) |
| Vegetated urban canopy model | Lee & Park (2007) |

**Table S3.** Morphological parameters provided for the four different Stages (1-4) in Grimmond et al. (2011). These parameters are used to configure TEB and WRF-TEB. For WRF-TEB the same parameters are used for all grid points classified as urban. Material characteristics provided in Stage 4 have information for four layers per facet (roof, wall, and road): composition/material, width ($d$, mm), volumetric heat capacity ($c$, MJ m$^{-3}$ K$^{-1}$), and thermal conductivity ($\lambda$, W m$^{-1}$ K$^{-1}$). Parameters not provided, but required to run TEB/WRF-TEB, are set to their default. [a]TEB/WRF-TEB: not used; [b]TEB/WRF-TEB: aggregated to a single vegetation value; [c]TEB/WRF-TEB: same value used for all wall/roof/road facets. *In Grimmond et al. (2011) modelers were not given the exact latitude and longitude to keep the site anonymous; here these are given for TEB and WRF-TEB simulations.

| Stage | Category | Data provided | | | | | | | | |
|---|---|---|---|---|---|---|---|---|---|---|
| 1 | Forcing | See Table 2a | | | | | | | | |
| | Site | *Lat. =-37.7306 °N, *Long. = 145.0145 °E; Measurement height = 30 m | | | | | | | | |
| 2 | Area fraction | Pervious = 0.38; Impervious = 0.62 | | | | | | | | |
| 3 | Heights | Instrument height = 40 m; Roughness length for momentum = 0.4 m; [a]Max height of roughness elements = 12 m; Mean building height = 6.4 m; Height to width ratio = 0.42; [a]Mean wall to plan area ratio = 0.4 | | | | | | | | |
| | Area fraction | Building = 0.445; Concrete = 0.045; Road = 0.130; [b]Vegetation (not Grass) = 0.225; [b]Grass = 0.150; [b]Other (bare or pools) = 0.005 | | | | | | | | |
| | Other | Urban climate zone = 5; Population density = 415.78 inhabitants km$^{-2}$ | | | | | | | | |
| 4 | Buildings | | Wall | | | Roof | | | Road | | |
| | | Layer | $d$ | $c$ | $k$ | $d$ | $c$ | $k$ | $d$ | $c$ | $k$ |
| | | 1 | 40.40 | 1.25 | 0.61 | 11.6 | 2.07 | 6.530 | 28.75 | 1.14 | 1.17 |
| | | 2 | 54.00 | 1.40 | 0.430 | 50.00 | 0.0071 | 0.025 | 158.30 | 1.05 | 0.30 |
| | | 3 | 42.00 | 0.0013 | 0.024 | 40.00 | 1.50 | 0.230 | 112.50 | 1.29 | 0.42 |
| | | 4 | 12.50 | 0.67 | 0.160 | 12.50 | 0.67 | 0.160 | 650.45 | 1.43 | 3.72 |
| | Surface | [c]Surface albedo = 0.15; [c]Surface emissivity = 0.97 | | | | | | | | |

**Table S4.** Main WPS/WRF configuration settings used with the model timestep for each domain (d1–d4). Lambert Conformal Conic (LCC). [†]Vertical grid spacing increasing with height (h) and first level (L1) set to 66 m a.g.l.

| Option | Value | TS/Unit | Reference |
|---|---|---|---|
| a) *Time* | | | |
| Timestep length | 135, 45, 15, 5 | s | - |
| b) *Grid* | | | |
| Map Projection | LCC | - | - |
| Horizontal Spacing | 27, 9, 3, 1 | km | - |
| Vertical Spacing | $f(h)$ with L1 = 66[†] | m | - |
| Vertical Levels | 61 | - | - |
| Nests and Grid Ratio | (2)4 and 1:3 | - | - |
| Nesting Approach | 1-way | - | - |
| Urban Classes | 1 | - | - |
| c) *Initial and Boundary Conditions* | | | |
| Data Set Name | ECMWF Cycle 28r2 analysis | - | ECMWF (2004) |
| Horizontal Spacing | TL511 ($\approx$ 40 km) | - | - |
| Vertical Levels | 61 | - | - |
| Time Interval | 6 | h | - |
| d) *Physical Parametrization* | | | |
| Shortwave Radiation | RRTMG | - | Iacono et al. (2008) |
| Longwave Radiation | RRTMG | - | Iacono et al. (2008) |
| Microphysics | Single-moment 3-class | - | Hong & Lim (2006) |
| Cumulus | New Tiedtke Scheme | - | Zhang & Wang (2017) |
| PBL | YSU | - | Hong et al. (2006) |
| Surface layer | Revised MM5 | - | Jiménez et al. (2012) |
| LSM | Noah-LSM | - | Chen & Dudhia (2001) |
| ULSM | TEB/UNN | - | Meyer, Schoetter, Riechert, et al. (2020)/This work |